\documentclass{article} 
\usepackage{iclr2027_conference,times}

\usepackage{amsmath,amsfonts,bm}

\def\eqref#1{equation~\ref{#1}}

\def\1{\bm{1}}

\DeclareMathAlphabet{\mathsfit}{\encodingdefault}{\sfdefault}{m}{sl}
\SetMathAlphabet{\mathsfit}{bold}{\encodingdefault}{\sfdefault}{bx}{n}

\usepackage[utf8]{inputenc}
\usepackage[T1]{fontenc}
\usepackage{url}
\usepackage{booktabs}
\usepackage{amsmath}
\usepackage{amsfonts}
\usepackage{nicefrac}
\usepackage{microtype}
\usepackage{xcolor}
\usepackage{graphicx}
\usepackage{multirow}
\usepackage{pifont}
\usepackage{colortbl}
\usepackage{threeparttable}
\usepackage{xspace}
\usepackage{todonotes}
\usepackage[etex=true,export]{adjustbox}
\usepackage{makecell}
\usepackage{subcaption}
\usepackage{epsfig}
\usepackage{epstopdf}
\usepackage{bbding}
\usepackage{diagbox}
\usepackage{footnote}
\usepackage{wrapfig}
\usepackage[normalem]{ulem}
\usepackage{enumitem}
\usepackage{tikz}
\usepackage{algorithm}
\usepackage{algpseudocode}
\usepackage{textcomp}
\usepackage{chngcntr}

\usepackage{tabularx}
\usepackage{xcolor}
\usepackage{colortbl}
\providecolor{memyellow}{HTML}{FFF9DB}
\newcommand{\ub}[1]{\textcolor[rgb]{.502,.502,.502}{#1}}

\usepackage[most]{tcolorbox}
\usepackage{fvextra}

\definecolor{promptblue}{HTML}{243B55}
\definecolor{promptfill}{HTML}{F7F9FC}

\newtcblisting{memprompt}[2][]{
enhanced,
listing only,
title={#2},
colback=promptfill,
colframe=promptblue,
colbacktitle=promptblue,
coltitle=white,
fonttitle=\bfseries\small,
arc=2pt,
boxrule=0.6pt,
left=6pt,
right=6pt,
top=4pt,
bottom=4pt,
listing options={
basicstyle=\ttfamily\scriptsize,
breaklines=true,
breakatwhitespace=false,
columns=fullflexible,
keepspaces=true,showstringspaces=false,
literate={—}{{\textemdash}}1
         {’}{{\textquoteright}}1
},
#1
}

\title{MemFolder: Structure-Guided Memory Folding for Efficient Agent Memory Management}
\author{Xuyang Liu$^{1}$\thanks{Equal contribution. $^\text{\Envelope}$Corresponding author.} \quad Chenwen Lu$^{1*}$ \quad Junxi Wang$^2$ \quad Siyu Ren$^1$ \quad Xinying Lin$^3$ \\
\textbf{Siteng Huang$^4$} \quad 
\textbf{Hongli Sun$^1$} \quad \textbf{Linfeng Zhang}$^5$ \quad 
\textbf{Honggang Chen}$^{1\text{\Envelope}}$\\
$^1$Sichuan University \quad 
$^2$Fudan University \quad
$^3$Sun Yat-sen University \\ 
$^4$Zhejiang University \quad 
$^5$Shanghai Jiao Tong University\\
\vspace{5pt}
\textbf{Code:} \href{https://github.com/xuyang-liu16/MemFolder}{\textcolor{citecolor}{\textbf{\texttt{https://github.com/xuyang-liu16/MemFolder}}}}
}

\definecolor{citecolor}{HTML}{0071bc}
\usepackage[colorlinks=true,linkcolor=red,citecolor=citecolor]{hyperref}
\usepackage[capitalise]{cleveref}

\iclrfinalcopy 
\begin{document}

\maketitle

\vspace{-20pt}
\begin{abstract}
\vspace{-5pt}
Long-term agents accumulate large external memory stores as interactions unfold, increasing the costs of storage, retrieval, and response generation.
Efficiency is constrained by two coupled factors: a growing record count increases retrieval overhead, while longer retrieved contexts increase prefill computation and decoding-time attention costs.
Existing methods prune or consolidate records, but pruning can discard useful evidence, whereas consolidation may produce fewer yet longer retrieval units.
We introduce \textbf{MemFolder}, a structure-guided memory folding framework that sequentially reduces record count and context length.
The framework comprises two core components that share memory groups, using cross-record structure to guide both evidence consolidation and compact representation.
\textbf{RecordFolder} grounds entity mentions, progressively selects representative yet complementary anchors, and combines semantic and entity affinity with temporal chaining to consolidate compatible records into fewer retrieval units.
\textbf{ContextFolder} encodes the original records within these groups into compact text and uses verification feedback to revise encodings when key facts cannot be recovered.
Experiments on the three text-based memory benchmarks LoCoMo, PersonaMem, and LongMemEval and the two multimodal benchmarks M3-Bench-web and M3-Bench-robot demonstrate the strong performance of our methods.
Notably, on Mem0, RecordFolder retains \textbf{99.0\%} of the original average performance across the three text-based benchmarks with 50\% fewer records, exceeding the strongest baseline by \textbf{1.9} percentage points in performance retention.
Combined with ContextFolder, existing record compression methods further shorten their output records while retaining most of their performance.
\end{abstract}
\section{Introduction}
\vspace{-5pt}

Advances in reasoning and tool use enable large language models~\citep{xu2026deepseek-v4,kimi2026k25,kimi2026k3} to support long-term agents operating across multiple sessions~\citep{wu2025longmemeval,he2026memoryarena}. These agents draw on earlier conversations, observations, and task states to inform later decisions. External memory systems~\citep{chhikara2025mem0,xu2025amem,liu2026simplemem} index this growing history as natural-language records for retrieval. Our uncompressed stores contain \textbf{6,817 records} totaling \textbf{61,273 tokens} across all ten LoCoMo samples~\citep{maharana2024locomo}, and \textbf{310,120 records} totaling \textbf{3.7M tokens} across LongMemEval~\citep{wu2025longmemeval}.

\begin{figure}[!t]
    \centering
    \includegraphics[width=\linewidth]{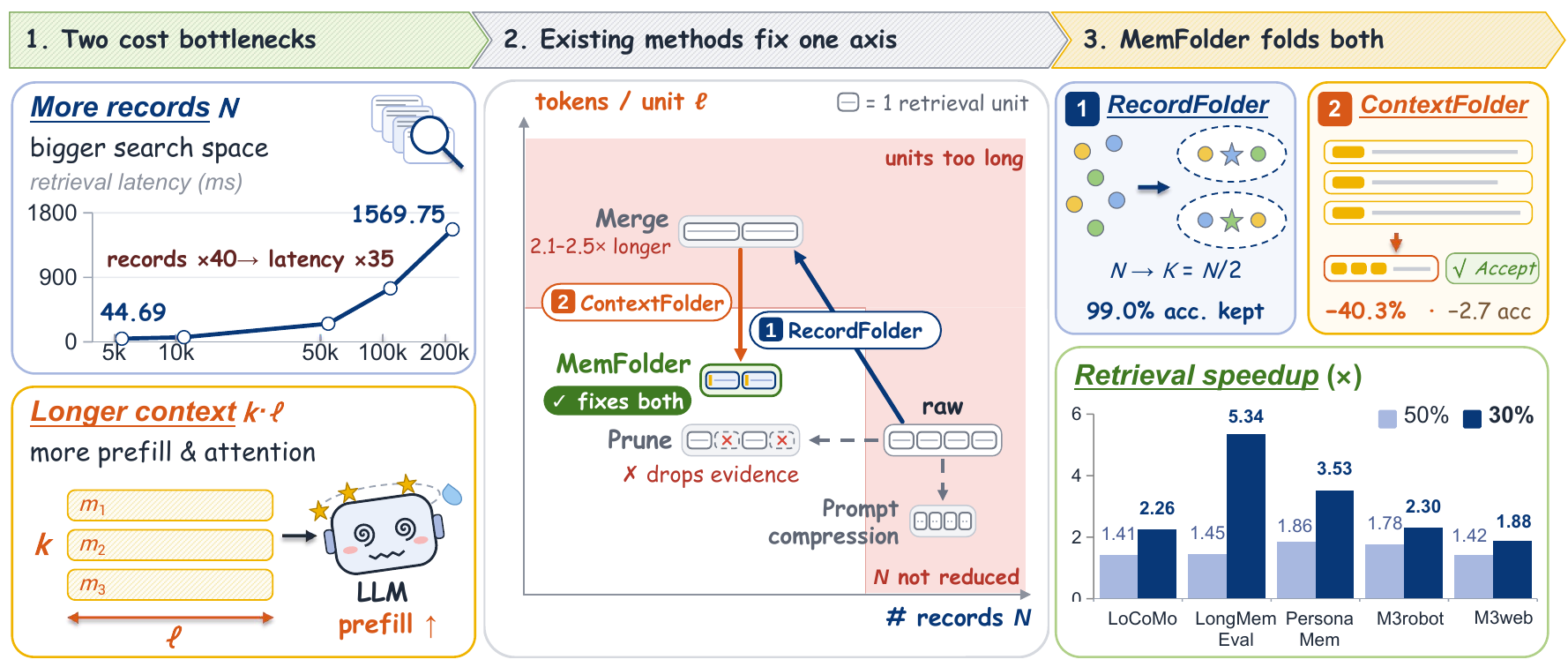}
    \vspace{-5mm}
    \caption{\textbf{MemFolder for efficient agent memory management.}
    \textbf{(1)} More records increase retrieval costs, while longer contexts increase prefill and token overhead.
    \textbf{(2)} Existing methods target one axis, while MemFolder combines RecordFolder and ContextFolder to reduce both.
    \textbf{(3)} MemFolder preserves accuracy while accelerating retrieval and reducing context length.}
    \label{fig:teaser}
    \vspace{-5mm}
\end{figure}

This growth creates \textbf{two coupled efficiency bottlenecks}, as shown in Figure~\ref{fig:teaser}(1). A larger \textbf{record count} requires more storage, increases the size of the index, and expands the retrieval search space. Once the records have been retrieved, their combined \textbf{context length} determines how much memory text the agent processes per query. Even individually short records can form a long context when retrieved together, increasing both prefilling and decoding costs~\citep{liu2025shifting}.

Figure~\ref{fig:teaser}(2) plots record count $N$ on the horizontal axis and tokens per unit $\ell$ on the vertical axis.
Prompt compressors such as LLMLingua~\citep{jiang2023llmlingua,pan2024llmlingua2} and BabelTele~\citep{zhu2026babeltele} shorten the input to each call while leaving $N$ unchanged.
StreamMeCo~\citep{wang2026streammeco} and MemForest~\citep{wang2026memforest} reduce $N$ by pruning or consolidating records. Pruning can discard useful evidence. Consolidation preserves related content but can produce fewer, longer units, shifting the store toward the upper left in the figure.
On LoCoMo~\citep{maharana2024locomo}, when 50\% of records are retained, units produced by MemForest and ToMe~\citep{bolya2023tome} average $2.12\times$ and $2.48\times$ the original record length, respectively. At 30\% retention, the ratios rise to $3.01\times$ and $3.87\times$.
Even concatenation without merging yields units $2.10\times$ the original length at 50\% retention.
With a fixed top-$k$ retrieval budget, \textbf{fewer units can still yield a longer context}. We aim to reduce both costs while retaining evidence an agent may need in later sessions.

This observation motivates two distinct compression decisions.
At the \textbf{record level}, we determine which records can share a retrieval unit while preserving complementary facts and state changes.
At the \textbf{context level}, we determine how to encode each group compactly, expressing shared content once while retaining dates and other record-specific details.
Group assignments connect these decisions while allowing record count and context length to be optimized separately.

We address these two efficiency bottlenecks with \textbf{MemFolder}, a structure-guided memory folding framework with two complementary folding components. As shown in Figure~\ref{fig:teaser}(3),
\textbf{RecordFolder} grounds entity mentions, progressively selects representative yet complementary anchors, and combines semantic and entity affinity with temporal chaining to consolidate $N$ records into $K$ retrieval units.
Given these groups and their source records, \textbf{ContextFolder} generates compact encodings guided by a target token ratio while keeping one output unit per group.
Verification feedback guides revisions when key facts are not recoverable.
Given that it is independent of the grouping algorithm, ContextFolder can also shorten the contexts from record compression methods.

Experiments cover three text benchmarks and two multimodal settings.
With 50\% fewer records, RecordFolder retains \textbf{99.0\%} of average Mem0 performance across the three text benchmarks, compared with 97.1\% for MemForest.
Combined with RecordFolder, ContextFolder further cuts tokens per record by 40\%, 38\%, and 31\% on LoCoMo, LongMemEval, and M3-Bench-robot, respectively, with limited accuracy loss.
As a plug-and-play component, ContextFolder also shortens outputs from other record-compression methods, yielding more favorable accuracy--efficiency trade-offs than BabelTele and LLMLingua-2.
Our work makes the following \textbf{contributions}:

\begin{enumerate}[leftmargin=12pt, topsep=2pt, itemsep=2pt]
    \item We introduce MemFolder, a structure-guided memory folding framework that jointly reduces the number of retrieval units and their context length for long-term agents.
    
    \item We develop RecordFolder to consolidate records through progressive anchor selection and entity-aware temporal chaining. ContextFolder then compacts each group's content through budgeted encoding and factual recovery checks that guide revisions to the encoding.
    
    \item Experiments on three text benchmarks and two multimodal settings demonstrate strong performance under record compression, while context compression analyses show further reductions in retrieval unit length across multiple existing memory consolidation methods.
\end{enumerate}

\section{Related Work}
\vspace{-5pt}

\noindent\textbf{LLM Agent Memory Systems.}
Memory systems preserve information from past interactions for use beyond the current context window.
Token-level external memory retains explicit records that can be inspected, updated, and retrieved without modifying model parameters.
Generative Agents~\citep{park2023generative} stores observations and reflections in a memory stream, while MemGPT~\citep{packer2023memgpt} manages information across memory tiers.
Mem0~\citep{chhikara2025mem0} extracts and consolidates salient facts from conversations, and A-MEM~\citep{xu2025amem} organizes structured notes through dynamically generated links.
LightMem~\citep{fang2026lightmem} organizes interactions across sensory, topic-aware short-term, and sleep-time consolidation stages, whereas REMem~\citep{shu2026remem} links semantically related episodic memories with compact gists and temporal facts to support iterative agentic retrieval.
Other approaches maintain experience in latent representations or model parameters.
MemoryLLM~\citep{wang2024memoryllm} uses a fixed-size, self-updatable latent memory pool, whereas TMEM~\citep{ren2026tmem} incorporates distilled experience through online updates to fast LoRA weights.
These forms can coexist: MemOS~\citep{li2025memos} unifies the management of plaintext, activation-based, and parameter-level memories.
Our work focuses on accumulated external records, whose count and textual content determine the size of the store and the evidence available for retrieval.
Rather than introducing another memory architecture, MemFolder compresses these records within existing systems while preserving text for retrieval and generation.

\noindent\textbf{Long-Term Memory Compression.}
Memory efficiency can be improved during construction, maintenance, or inference.
MemoryBank~\citep{zhong2024memorybank} updates retention through forgetting and reinforcement, while SimpleMem~\citep{liu2026simplemem} constructs compact memory units and recursively consolidates related content.
At inference time, LLMLingua~\citep{jiang2023llmlingua} shortens prompts, whereas token-level compression methods~\citep{li2024snapkv,liu2026mixkv,liu2026globalcom2,liu2025vidcom2,lin2026v-cast} reduce the states retained during generation.
These inference-side interventions do not by themselves reduce the number of records in a persistent external store.
We study post-construction compression before retrieval, so compressed records can be reused across queries.

Post-construction methods reduce existing stores through pruning or consolidation.
StreamMeCo~\citep{wang2026streammeco} prunes memory graphs using representative sampling and scores based on entity importance and semantic diversity, without rewriting the retained records.
MemRefine~\citep{kim2026memrefine} uses embedding similarity to propose candidate pairs, then an LLM judge determines whether their factual content warrants deletion, merging, or preservation.
Its refinement loop targets a storage budget measured over serialized entries.
MemForest~\citep{wang2026memforest} combines global semantic similarity with local temporal continuity to partition memories into EventTrees, then progressively merges records along selected tree edges.
It also introduces anchor-guided propagation retrieval to supplement evidence with temporally related memories.

MemFolder targets both record count and per-unit context length.
Unlike end-to-end memory systems~\citep{liu2026simplemem,fang2026lightmem,shu2026remem}, which either merge records without bounding per-unit token expansion or expand stores with multi-granularity graph nodes and multi-step retrieval, MemFolder operates as a plug-and-play post-construction layer that jointly folds record count and per-unit length over existing stores.
Compared with EventTree-based merging~\citep{wang2026memforest}, RecordFolder groups records around progressively selected anchors and refines assignments through entity-aware temporal chains.
ContextFolder reuses these groups to generate compact encodings under a soft token budget.
It checks whether source-derived facts can be recovered from the encoding without exposing the original records to the reader, and uses the feedback for revision.
This post-encoding check differs from deciding whether a candidate pair should be merged: it evaluates factual recovery from the resulting compact representation.
\section{Methodology}
\vspace{-5pt}

\subsection{Preliminary: Agent Memory Systems and Compression}
\noindent \textbf{Agent Memory Systems.}
An agent memory system converts an interaction history $\mathcal{H}_t$ into a persistent external store $\mathcal{M}_t=\{m_i\}_{i=1}^{N_t}$, where $N_t$ is the number of stored records.
When a new interaction $x_t$ arrives, an extractor $\mathcal{E}$ produces candidate records $\Omega_t=\mathcal{E}(x_t,\mathcal{H}_{<t})$, and an updater $\mathcal{U}$ incorporates them as $\mathcal{M}_t=\mathcal{U}(\mathcal{M}_{t-1},\Omega_t)$.
Given a query $q$, retrieval returns $\mathcal{R}_k(q,\mathcal{M}_t)=\operatorname{TopK}_{m\in\mathcal{M}_t}s(q,m)$, where $s(q,m)$ measures relevance.
The agent conditions its response on $q$ and the retrieved records.
As interactions accumulate, the store requires more persistent storage and exposes more candidates to each retrieval operation as the store grows.

\noindent \textbf{Memory Compression.}
Accumulated agent memories contain redundancy at two levels: redundant \textbf{records} across the store and repeated \textbf{context} within each record.
For a fixed snapshot $\mathcal{M}=\mathcal{M}_t$, let $N=N(\mathcal{M})=|\mathcal{M}|$ denote its record count.
Record compression reduces this count by retaining representative records or consolidating related ones.
Given a record keep ratio $\kappa\in(0,1]$, it targets $K=\max\{1,\operatorname{round}(\kappa N)\}$ retrieval units.
Context compression reduces the total token length $L(\mathcal{M})=\sum_{m\in\mathcal{M}}\ell(m)$, where $\ell(m)$ denotes the token length of record $m$, by expressing shared facts once while retaining group-specific details within each retrieval unit.

\subsection{MemFolder Overview}

\begin{figure}[!t]
    \centering
    \includegraphics[width=\linewidth]{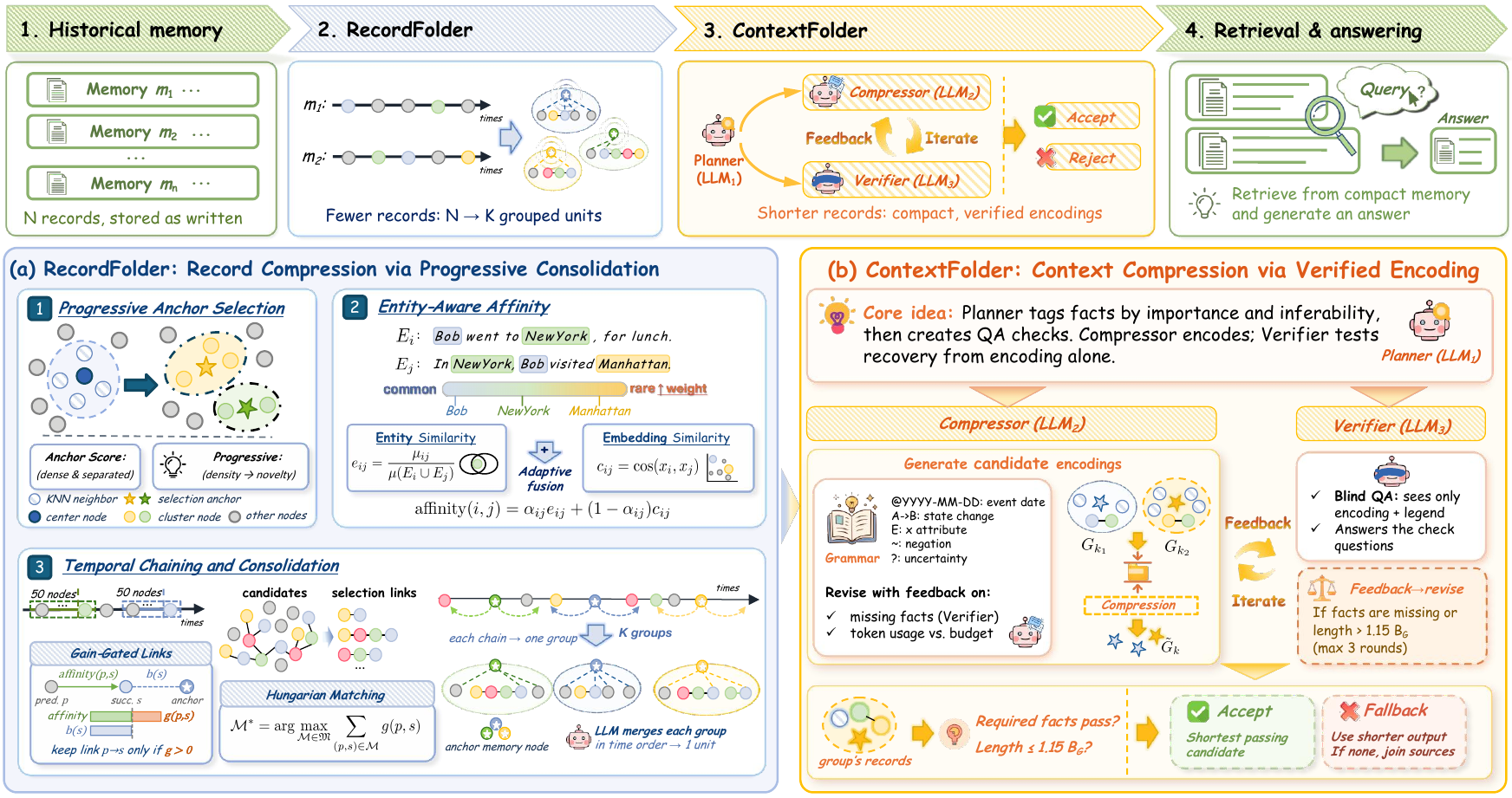}
    \vspace{-5mm}
    \caption{\textbf{Overview of MemFolder.}
    \textbf{(a) RecordFolder} maps $N$ records to $K$ retrieval units through entity grounding, complementary anchor selection, entity-aware affinity, and Hungarian matching of temporal links.
    \textbf{(b) ContextFolder} compresses each group's source records to a target token-retention ratio using a planner that prioritizes facts and prepares checks, a compressor that generates grammar-based candidates, and a blind verifier that tests recovery over up to three rounds.
    ContextFolder keeps the shortest candidate passing required checks; otherwise, weighted recovery guides selection, with a shorter candidate or source texts used when compression would expand the memory.}
    \label{fig:overview}
    \vspace{-5mm}
\end{figure}

Figure~\ref{fig:overview} presents the pipelines of our MemFolder, which addresses both memory costs by sequentially folding the record store along two axes. \textbf{RecordFolder} grounds records, selects representative and complementary anchors, refines group assignments using entity, semantic, and temporal relations, and consolidates each group into one retrieval unit. \textbf{ContextFolder} reuses these groups to generate compact grammar-based encodings under group-specific budgets: a planner identifies facts to preserve, a compressor rewrites each group, and a verifier checks recovery from the encoding alone. Thus, retrieval accesses fewer units with less context in each.

\subsection{RecordFolder: Record Compression via Progressive Consolidation}
Records that refer to the same entities or events can remain in separate retrieval units, splitting related evidence across results.
\textbf{RecordFolder} maps $N$ records to $K$ units by grounding entities, selecting complementary anchors, and refining assignments with entity-aware affinity and temporal links.
Hungarian matching restricts these links to non-branching chains. An LLM then consolidates each multi-record group chronologically, while singleton records remain unchanged.

\noindent \textbf{Entity Grounding.}
The same entity may appear under different names or pronouns, so surface overlap alone can miss connections between records.
We use an LLM to extract entities, including people, works, activities, and topics, and assign short canonical labels.
Within each session, we process records together to resolve references in context, then process sessions chronologically while carrying earlier labels forward in a registry.
The prompt reuses labels for known referents and adds qualifiers to distinguish namesakes.
We merge residual label variants whose embedding cosine similarity is at least $\theta_{\mathrm{ent}}$, retaining one representative per connected component.
After removing conversation speakers and synthetic event identifiers, the remaining labels form the entity set $E_i$ of record $m_i$ for affinity computation and later temporal refinement.

\noindent \textbf{Progressive Anchor Selection.}
Anchors define the $K$ groups that become retrieval units.
At each step, we score every remaining record against the anchors already selected.
Let $x_i$ be the normalized embedding of record $m_i$ and $D_{ij}=\|x_i-x_j\|_2^2$.
We select the first anchor with the density-peak clustering criterion, DPC~\citep{rodriguez2014dpc}.
We compute local density $\rho_i=\exp(-\frac{1}{k}\sum_{j\in\operatorname{kNN}(i)}D_{ij})$ from the $k=\min(k_{\mathrm{nn}},N-1)$ nearest other records, where $k_{\mathrm{nn}}$ caps the neighborhood size.
Separation is $\delta_i=\min_{j:\rho_j>\rho_i}D_{ij}$, the distance to the nearest record with higher density.
We choose $a_1=\arg\max_i\rho_i\delta_i$, favoring a locally dense record separated from denser alternatives.
Subsequent selections replace local density with novelty relative to the set $A_r$ of $r$ selected anchors:
\begin{equation}
\rho_i^{(r)}=1-\exp\Bigl(-\frac{1}{k'}\sum_{a\in\operatorname{kNN}_{A_r}(i)}D_{ia}\Bigr),\qquad
\delta_i^{(r)}=\min_{\substack{j\notin A_r\\\rho_j^{(r)}>\rho_i^{(r)}}}D_{ij},
\end{equation}
where $k'=\min(k_{\mathrm{nn}},|A_r|)$ and $\operatorname{kNN}_{A_r}(i)$ contains the $k'$ nearest selected anchors.
When no unselected record has a higher $\rho$ value, we set $\delta_i^{(r)}$ to the maximum distance from $i$ to an unselected record.
We select $a_{r+1}=\arg\max_{i\notin A_r}\rho_i^{(r)}\delta_i^{(r)}$ until $K$ anchors are chosen.
The novelty term favors candidates farther from existing anchors, encouraging complementary coverage.
Nearest-anchor assignment yields the initial partition $\pi_0$, with $a_0(i)$ denoting the anchor of $m_i$.

\noindent \textbf{Entity-Aware Affinity.}
The initial partition uses embeddings; entity overlap adds a complementary signal when linking records over time.
To downweight common entities, we assign $w(e)=\log(N/\operatorname{df}(e))$, where $\operatorname{df}(e)$ counts the records containing entity $e$.
For an entity set $S$, let $\mu(S)=\sum_{e\in S}w(e)$ denote its total weight.
The shared entity mass is $\mu_{ij}=\mu(E_i\cap E_j)$, and the entity similarity is $e_{ij}=\mu_{ij}/\mu(E_i\cup E_j)$, set to zero when the denominator is zero.
We combine entity similarity with cosine similarity $c_{ij}=\cos(x_i,x_j)$ as $\operatorname{affinity}(i,j)=\alpha_{ij}e_{ij}+(1-\alpha_{ij})c_{ij}$.
The mixing weight is $\alpha_{ij}=\mu_{ij}\sigma_c^2/(\mu_{ij}\sigma_c^2+\sigma_e^2)$.
We compute signal variances $\sigma_e^2$ and $\sigma_c^2$ over all earlier--later pairs in a $W$-record window before filtering, using a positive floor for numerical stability.
The affinity upweights rare shared entities and equals cosine similarity when $\mu_{ij}=0$.

\noindent \textbf{Temporal Chaining and Consolidation.}
We use this affinity to refine the initial assignments by linking records that may continue the same event.
For each non-anchor record $m_s$, we inspect up to $W$ predecessors $m_p$ and retain at most $d$ links with positive gain $g(p,s)=\operatorname{affinity}(p,s)-\cos(x_s,x_{a_0(s)})$.
Thus, a predecessor is retained only when its affinity with $m_s$ exceeds the cosine similarity between $m_s$ and its initial anchor.
Hungarian matching maximizes total gain between predecessor and successor copies, allowing each record at most one incoming and one outgoing edge.
Because every link follows temporal order, the matched links form non-branching chains.
Anchors have no incoming links, so a chain containing an anchor begins with it and remains in its group.
We assign each anchor-free chain $C$ to $a^\star(C)=\arg\max_{a\in A_K}\sum_{i\in C}\cos(x_i,x_a)$ and move its members together.
Unmatched records retain their initial assignments in $\pi_0$, preserving the $K$ groups.
An LLM consolidates each multi-record group in one call, reading members chronologically and resolving relative dates from their timestamps.
Singleton records remain unchanged, and we embed all resulting units for downstream retrieval and later query answering.

\subsection{ContextFolder: Context Compression via Verified Encoding}
Consolidating records can reduce the number of retrieval units while lengthening each one; at a fixed retrieval top-$k$, this need not reduce the context passed to the agent.
\textbf{ContextFolder} compresses the original records within each assigned group guided by a target fraction of source tokens, without changing group membership.
A planner labels facts by importance and inferability and prepares QA checks. A compressor generates candidate encodings, and a blind verifier tests whether a reader can recover the facts from the encoding alone.
Feedback on missing facts and token usage guides revision while preserving each group's assignment and target ratio throughout compression.

\noindent \textbf{Planner: Fact Extraction and Triage.}
For each group $G$, an LLM extracts atomic facts from its member records and writes a verification question with an expected answer for each fact.
These records are the planner's only source; downstream test queries and answers do not enter the process.
The planner labels facts as high, medium, or low importance, prioritizing key events, entity bindings, exact dates and numbers, and negations.
It also marks facts inferable from other facts or common knowledge.
We weight facts by importance and discount inferable facts when computing weighted recovery.
Let $\tau\in(0,1)$ denote the target fraction of source tokens to retain.
Non-inferable high-importance facts are always required.
Medium-importance facts are also required when non-inferable and $\tau\geq\tau_{\mathrm{req}}$, where $\tau_{\mathrm{req}}$ is a verification threshold.

\noindent \textbf{Compressor: Ratio-Guided Grammar Encoding.}
For a group $G$, the compressor uses budget $B_G=\max\{3,\operatorname{round}(\tau S_G)\}$, where $S_G=\sum_{m\in G}\ell(m)$ is its total source length.
The same $\tau$ therefore sets a proportional budget for each group, not a fixed token count.
The encoding uses \texttt{@YYYY-MM-DD} for event dates, \texttt{A -> B} for state changes, and \texttt{E: x} for attributes.
The symbols \texttt{\textasciitilde} and \texttt{?} indicate negation and uncertainty.
The prompt prioritizes non-inferable, high-importance facts, omitting inferable details first and lower-priority content as needed.
For retained facts, it requests exact entity names, numbers, and dates.
Both the verifier and downstream agent use the same grammar legend to decode these compact symbols consistently.

\noindent \textbf{Verifier: Blind Recovery Check.}
Separate answering and judging calls evaluate each candidate.
A blind reader receives the encoding, grammar legend, and verification questions, but not the source records or expected answers.
It returns \texttt{UNKNOWN} when the encoding lacks sufficient evidence.
A judge compares the responses with the reference facts, allowing semantic matches while requiring exact names, dates, and numbers.
We run at most $R_{\max}$ encoding-verification rounds.
The loop stops when all required facts pass verification and the encoding uses at most $(1+\epsilon_B)B_G$ tokens, where $\epsilon_B$ is the budget tolerance.
Otherwise, feedback guides revision until the iteration limit.
We select the shortest candidate recovering all required facts.
If none passes, we maximize weighted recovery, breaking ties by length.
Weighted recovery is the fraction of total fact weight judged recoverable.
If the selected encoding is no shorter than $S_G$, we use the shortest candidate below $S_G$, even if it misses required facts. If no candidate is shorter, we retain the source texts as one unit.
Because the compressor conditions on both the prioritized fact list and the raw source records (Appendix~\ref{app:prompts}), details omitted during triage remain accessible, while the blind verification loop acts as an asymmetric failure-detection heuristic that guards against silent loss of required names, dates, and state changes (see Appendix~\ref{app:check-noise} for reliability traces).


\section{Experiments}
\vspace{-5pt}
\label{sec:experiments}

\subsection{Experimental Setting}
\label{subsec:experimental_setups}

\noindent \textbf{Benchmarks.}
We evaluate MemFolder on four benchmarks: LoCoMo~\citep{maharana2024locomo}, PersonaMem~\citep{jiang2025personamem}, LongMemEval~\citep{wu2025longmemeval}, and M3-Bench~\citep{long2026m3-agent}, covering its web and robot subsets.
For each benchmark and memory framework, all methods start from the same store constructed from the original interaction histories.


\noindent \textbf{Implementations.}
     All experiments are conducted on two NVIDIA A100 (80GB) GPUs.
     We use \texttt{gpt-4o-mini} for memory node merging.
     In addition, all other settings, including those of the compared baselines and
     the underlying memory frameworks, follow the original papers.
     All experimental results are averaged over three runs.
     More details are provided in Appendix~\ref{app:additional-details}.

\noindent \textbf{Baselines.}
For record compression, we compare with pruning baselines (Random Pruning, KMeans, and DART~\citep{wen2025dart}) and merging baselines (Random Merging, ToMe~\citep{bolya2023tome}, and MemForest~\citep{wang2026memforest}).
For context compression, we compare ContextFolder with LLMLingua-2~\citep{pan2024llmlingua2} and BabelTele~\citep{zhu2026babeltele}.

\subsection{Main Results}
\label{subsec:main_comparison}

\begin{table*}[t]
    \caption{\textbf{Record compression on Mem0 cross compression ratios.}
    \textbf{All}: overall accuracy; \textbf{Avg.}: mean accuracy across three benchmarks; \textbf{\%}: Avg.\ retained relative to uncompressed Mem0.
    Best and second-best compressed results per column and ratio are \textbf{bold} and \uline{underlined}, respectively.}
    \vspace{-2mm}
    \label{tab:main_mem0}
    \centering
    \setlength{\tabcolsep}{3pt}
    \renewcommand{\arraystretch}{1.05}
    \resizebox{\textwidth}{!}{
    \begin{tabular}{lccccc|ccccccc|c|c|c}
    \toprule
    \multirow{2}{*}[-0.8ex]{\textbf{Method}}
         & \multicolumn{5}{c|}{\textbf{LoCoMo}}
         & \multicolumn{7}{c|}{\textbf{LongMemEval}}
         & \multirow{2}{*}[-0.8ex]{\shortstack[c]{\textbf{Persona}\\\textbf{-Mem}}}
         & \multirow{2}{*}[-0.8ex]{\textbf{Avg.}}
         & \multirow{2}{*}[-0.8ex]{\textbf{\%}} \\
    \cmidrule(lr){2-6}\cmidrule(lr){7-13}
         & SH & MH & TR & OD & \textbf{All}
         & SU & SA & SP & MS & KU & TR & \textbf{All}
         &  &  &  \\
    \midrule
    \ub{Mem0}
         & \ub{66.0} & \ub{56.0} & \ub{53.9} & \ub{42.7} & \ub{60.2}
         & \ub{90.0} & \ub{21.4} & \ub{23.3} & \ub{54.9} & \ub{79.5} & \ub{44.4} & \ub{55.2}
         & \ub{67.9} & \ub{61.1} & \ub{100.0} \\
    \midrule
    \multicolumn{16}{l}{\textit{Record-count reduction: $30\%$}} \\
    \hspace{7pt}+ Random Pruning & 55.3 & 49.7 & 43.3 & 39.6 & 50.8 & 82.9 & 17.9 & \uline{26.7} & \uline{51.9} & 73.1 & 39.1 & 50.8 & 64.5 & 55.4 & 90.7 \\
    \hspace{7pt}+ KMeans & 61.1 & 54.3 & \uline{52.7} & 40.6 & 56.8 & 84.3 & 17.9 & 20.0 & \uline{51.9} & \uline{75.6} & 41.4 & 52.0 & 65.7 & 58.2 & 95.3 \\
    \hspace{7pt}+ DART & \uline{64.1} & 53.9 & 48.0 & 40.6 & 57.4 & 82.9 & 19.0 & 20.0 & 48.9 & 74.4 & 34.6 & 48.8 & \uline{66.6} & 57.6 & 94.3 \\
    \hspace{7pt}+ Random Merging & 60.9 & 52.1 & 51.1 & 43.8 & 56.2 & \textbf{91.4} & \textbf{21.4} & 20.2 & 46.6 & 74.4 & 41.4 & 50.8 & 65.2 & 57.4 & 93.9 \\
    \hspace{7pt}+ ToMe & 63.1 & 51.1 & 50.8 & \uline{44.8} & 57.2 & 85.7 & \textbf{21.4} & 23.3 & 48.1 & 71.8 & 42.9 & 51.2 & 66.4 & 58.3 & 95.4 \\
    \hspace{7pt}+ MemForest & 63.7 & \uline{54.6} & 52.0 & \textbf{45.8} & \uline{58.5} & \uline{90.0} & \uline{19.6} & 20.0 & 49.8 & \textbf{76.9} & \textbf{49.6} & \uline{54.2} & \uline{66.6} & \uline{59.7} & \uline{97.7} \\
    \rowcolor{memyellow}\hspace{7pt}+ \textbf{RecordFolder (Ours)} & \textbf{67.5} & \textbf{56.0} & \textbf{56.1} & 41.7 & \textbf{61.7} & 88.6 & \textbf{21.4} & \textbf{36.7} & \textbf{52.6} & \uline{75.6} & \uline{46.6} & \textbf{55.2} & \textbf{67.7} & \textbf{61.5} & \textbf{100.7} \\
    \midrule
    \multicolumn{16}{l}{\textit{Record-count reduction: $50\%$}} \\
    \hspace{7pt}+ Random Pruning & 44.4 & 45.4 & 37.1 & \textbf{44.8} & 43.1 & 75.7 & 17.9 & \uline{26.7} & 32.3 & 65.4 & 39.8 & 43.6 & 64.3 & 50.3 & 82.3 \\
    \hspace{7pt}+ KMeans & 51.6 & 43.6 & 39.6 & 37.5 & 46.8 & 77.1 & 19.6 & 23.3 & 39.1 & \uline{74.4} & 39.8 & 47.0 & 65.2 & 53.0 & 86.7 \\
    \hspace{7pt}+ DART & 57.4 & 49.6 & 41.7 & \uline{43.8} & 51.9 & 67.1 & 12.5 & 23.3 & 42.9 & 60.3 & 31.6 & 41.4 & 64.3 & 52.5 & 85.9 \\
    \hspace{7pt}+ Random Merging & 56.0 & 48.2 & 46.4 & 37.5 & 51.4 & 85.7 & \textbf{23.2} & 20.0 & 42.9 & 70.5 & 39.1 & 48.6 & 64.5 & 54.8 & 89.7 \\
    \hspace{7pt}+ ToMe & 60.2 & 50.4 & 47.4 & \uline{43.8} & 54.7 & 85.7 & \uline{21.4} & 23.3 & \uline{48.1} & 71.8 & 41.4 & 50.8 & 65.9 & 57.1 & 93.5 \\
    \hspace{7pt}+ MemForest & \uline{62.1} & \textbf{56.7} & \textbf{54.8} & \uline{43.8} & \uline{58.4} & \textbf{90.0} & 19.6 & 20.0 & \uline{48.1} & \uline{74.4} & \uline{42.9} & \uline{51.8} & \textbf{67.7} & \uline{59.3} & \uline{97.1} \\
    \rowcolor{memyellow}\hspace{7pt}+ \textbf{RecordFolder (Ours)} & \textbf{65.5} & \uline{55.7} & \uline{50.5} & 39.6 & \textbf{59.3} & \uline{88.6} & \uline{21.4} & \textbf{30.0} & \textbf{53.4} & \textbf{76.9} & \textbf{45.9} & \textbf{55.0} & \uline{67.2} & \textbf{60.5} & \textbf{99.0} \\
    \midrule
    \multicolumn{16}{l}{\textit{Record-count reduction: $70\%$}} \\
    \hspace{7pt}+ Random Pruning & 35.0 & 36.5 & 25.9 & 31.3 & 33.1 & 47.1 & 7.1 & \uline{30.0} & 16.5 & 42.3 & 31.6 & 28.6 & 63.7 & 41.8 & 68.4 \\
    \hspace{7pt}+ KMeans & 37.9 & 42.6 & 27.4 & 39.6 & 36.7 & 61.4 & 8.9 & \uline{30.0} & 21.8 & 59.0 & 31.6 & 34.8 & 63.8 & 45.1 & 73.8 \\
    \hspace{7pt}+ DART & 46.4 & 44.3 & 36.1 & 41.7 & 43.6 & 44.3 & 10.7 & 16.7 & 21.0 & 43.6 & 27.8 & 30.6 & 61.8 & 45.3 & 74.1 \\
    \hspace{7pt}+ Random Merging & 54.0 & 47.2 & 47.0 & \uline{43.8} & 50.7 & 80.0 & 21.4 & \uline{30.0} & 41.4 & 67.9 & 34.6 & 46.2 & 64.0 & 53.6 & 87.7 \\
    \hspace{7pt}+ ToMe & 56.6 & 51.8 & \uline{47.7} & 38.5 & 52.7 & \uline{84.3} & 21.4 & 20.0 & 45.1 & 70.5 & \uline{42.1} & 49.6 & \textbf{65.7} & 56.0 & 91.7 \\
    \hspace{7pt}+ MemForest & \uline{58.1} & \uline{56.4} & 45.2 & \textbf{45.8} & \uline{54.4} & \uline{84.3} & \textbf{25.0} & 23.3 & \textbf{52.6} & \uline{74.4} & 39.8 & \uline{52.2} & 64.4 & \uline{57.0} & \uline{93.3} \\
    \rowcolor{memyellow}\hspace{7pt}+ \textbf{RecordFolder (Ours)} & \textbf{61.2} & \textbf{57.0} & \textbf{50.7} & 43.4 & \textbf{57.1} & \textbf{85.7} & \uline{23.2} & \textbf{53.3} & \uline{50.4} & \textbf{75.6} & \textbf{45.1} & \textbf{55.0} & \uline{64.9} & \textbf{59.0} & \textbf{96.6} \\

    \bottomrule
    \end{tabular}
    }
\end{table*}

\begin{figure}[!t]
    \centering
    \includegraphics[width=\linewidth]{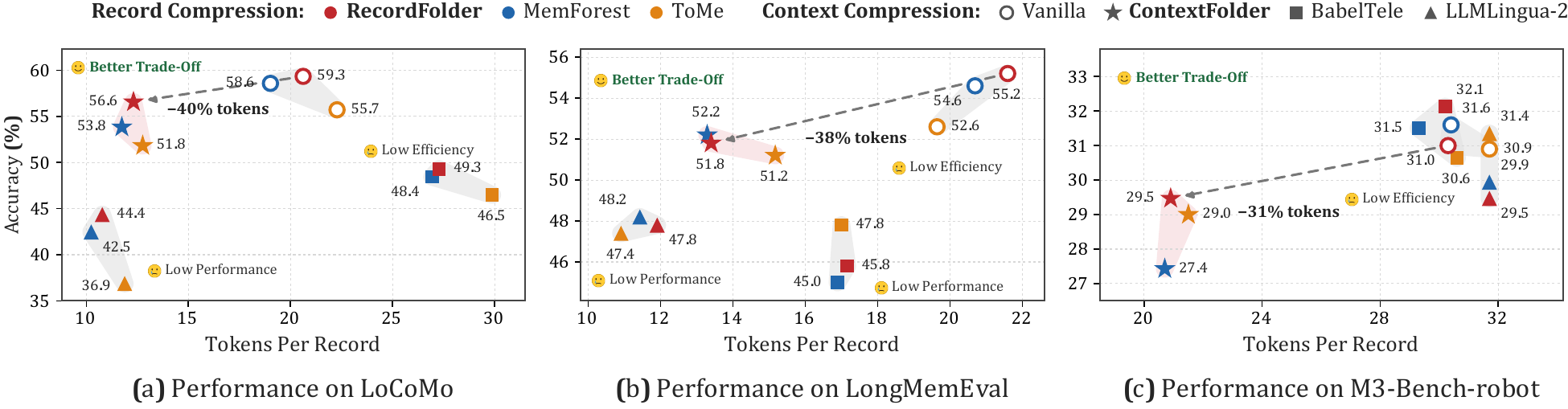}
    \vspace{-5mm}
    \caption{\textbf{Context compression results.}
    Accuracy--tokens trade-offs at 50\% record-count reduction.
    Colors and markers denote record and context compression methods.
    Upper left is better.}
    \label{fig:context_results}
    \vspace{-5mm}
\end{figure}

\noindent \textbf{Main Comparisons.}
Table~\ref{tab:main_mem0} compares RecordFolder with record-compression baselines on Mem0 at 30\%, 50\%, and 70\% record-count reduction.
Figure~\ref{fig:context_results}(a) and (b) compare context-compression methods after three record compressors at 50\% reduction.
\textbf{(i) RecordFolder leads at every ratio.}
It achieves the highest mean accuracy across the three benchmarks at every tested ratio.
At 50\% reduction, it retains \textbf{99.0\%} of original performance, compared with 97.1\% for MemForest.
At 70\%, these values are \textbf{96.6\%} and 93.3\%, respectively.
\textbf{(ii) Consolidation benefits LongMemEval.}
At 50\% reduction, RecordFolder scores 55.0 on LongMemEval, versus 51.8 for MemForest and 50.8 for ToMe.
Its 3.2-point lead over MemForest exceeds the 0.9-point gap on LoCoMo.
At 70\% reduction, it maintains 55.0, compared with 52.2 for MemForest and 49.6 for ToMe.
These results suggest that RecordFolder is particularly effective at preserving useful evidence when compressing long interaction histories.
\textbf{(iii) ContextFolder improves cross-method trade-offs.}
On LoCoMo, BabelTele increases tokens per record after all three record compressors.
After RecordFolder, tokens rise from 20.6 to 27.3 while accuracy drops from 59.3\% to 49.3\%.
This may reflect tokenization overhead from multilingual and symbolic rewriting of compact records.
On LongMemEval, BabelTele shortens RecordFolder's output, but ContextFolder achieves higher accuracy with fewer tokens: 51.8\% at 13.4 tokens per record, versus 45.8\% at 17.2.
ContextFolder also shortens MemForest and ToMe outputs on both benchmarks.
After RecordFolder, it exceeds LLMLingua-2 by 12.2 and 4.0 accuracy points on LoCoMo and LongMemEval, respectively, using about 1.5 additional tokens per record.
After RecordFolder halves the record count, ContextFolder further reduces tokens per record by \textbf{40.3\%} and \textbf{37.9\%} on the two benchmarks, respectively.

 \begin{table*}[t]
           \caption{
           \textbf{Record compression on MemOS across compression ratios.}
           }
           \vspace{-2mm}
           \label{tab:main_memos}
           \centering
           \setlength{\tabcolsep}{3pt}
           \renewcommand{\arraystretch}{1.05}
           \resizebox{\textwidth}{!}{
           \begin{tabular}{lccccc|ccccccc|c|c|c}
           \toprule
           \multirow{2}{*}[-0.8ex]{\textbf{Method}}
                & \multicolumn{5}{c|}{\textbf{LoCoMo}}
                & \multicolumn{7}{c|}{\textbf{LongMemEval}}
                & \multirow{2}{*}[-0.8ex]{\shortstack[c]{\textbf{Persona}\\\textbf{-Mem}}}
                & \multirow{2}{*}[-0.8ex]{\textbf{Avg.}}
                & \multirow{2}{*}[-0.8ex]{\textbf{\%}} \\
           \cmidrule(lr){2-6}\cmidrule(lr){7-13}
                & SH & MH & TR & OD & \textbf{All}
                & SU & SA & SP & MS & KU & TR & \textbf{All}
                &  &  &  \\
           \midrule

           \ub{MemOS}
                & \ub{84.5} & \ub{70.2} & \ub{78.8} & \ub{56.2} & \ub{79.0}
                & \ub{91.4} & \ub{100.0} & \ub{66.7} & \ub{55.6} & \ub{78.2} & \ub{71.4} & \ub{74.0}
                & \ub{60.3} & \ub{71.1} & \ub{100.0} \\

           \midrule
           \multicolumn{16}{l}{\textit{Record compression ratio $\downarrow 50\%$}} \\

           \hspace{7pt}+ Random Pruning
                & 70.5 & 65.2 & 57.3 & 46.9 & 65.3
                & 67.1 & 64.3 & 63.3 & 28.6 & 52.6 & 43.6 & 47.8
                & 55.9 & 56.3 & 79.2 \\
           \hspace{7pt}+ KMeans
                & 75.7 & 59.2 & 57.6 & \uline{57.3} & 67.8
                & 72.9 & 69.6 & 66.7 & 20.3 & 51.3 & 45.9 & 47.6
                & 54.7 & 56.7 & 79.7 \\
           \hspace{7pt}+ DART
                & 76.5 & 68.1 & 58.9 & \textbf{58.3} & 70.1
                & 62.9 & \textbf{82.1} & 50.0 & 20.3 & 35.9 & 33.1 & 40.8
                & 53.7 & 54.9 & 77.2 \\
           \hspace{7pt}+ Random Merging
                & 79.4 & \uline{69.9} & 66.0 & 54.2 & 73.3
                & 80.0 & 62.5 & \textbf{83.3} & \textbf{46.6} & 60.3 & 54.9 & 59.6
                & \uline{56.2} & 63.0 & 88.7 \\
           \hspace{7pt}+ ToMe
                & 76.7 & 68.8 & 63.9 & 52.1 & 71.0
                & 78.6 & 67.9 & \uline{76.7} & 40.6 & \uline{70.5} & \textbf{63.2} & \uline{61.8}
                & \textbf{57.9} & 63.6 & 89.4 \\
           \hspace{7pt}+ MemForest
                & \textbf{82.6} & 68.8 & \textbf{70.1} & \textbf{58.3} & \textbf{76.0}
                & \uline{81.4} & 76.8 & 66.7 & 40.6 & 65.4 & \uline{56.4} & 60.0
                & 54.8 & \uline{63.6} & \uline{89.5} \\
           \rowcolor{memyellow}
           \hspace{7pt}+ \textbf{RecordFolder (Ours)}
                & \uline{81.2} & \textbf{70.6} & \uline{69.5} & 56.2 & \uline{75.3}
                & \textbf{85.7} & \uline{78.6} & \textbf{83.3} & \uline{45.1} & \textbf{75.6} & 54.9 & \textbf{64.2}
                & 54.7 & \textbf{64.7} & \textbf{91.0} \\

           \midrule
           \multicolumn{16}{l}{\textit{Record compression ratio $\downarrow 70\%$}} \\

           \hspace{7pt}+ Random Pruning
                & 58.5 & 59.9 & 43.0 & 49.0 & 54.9
                & 32.9 & 48.2 & 73.3 & 13.5 & 29.5 & 28.6 & 30.2
                & 52.8 & 46.0 & 64.7 \\
           \hspace{7pt}+ KMeans
                & 62.5 & 57.8 & 43.9 & 55.2 & 57.3
                & 40.0 & 53.6 & 50.0 & 15.8 & 30.8 & 30.1 & 31.6
                & 52.6 & 47.2 & 66.4 \\
           \hspace{7pt}+ DART
                & 65.6 & 59.9 & 46.1 & 54.2 & 59.8
                & 30.0 & 60.7 & 50.0 & 15.0 & 25.6 & 18.0 & 26.8
                & 48.9 & 45.2 & 63.5 \\
           \hspace{7pt}+ Random Merging
                & 73.0 & 64.2 & \uline{61.7} & \uline{57.3} & 68.1
                & \textbf{75.7} & 39.3 & \uline{83.3} & \textbf{39.1} & 60.3 & \textbf{52.6} & \uline{53.8}
                & 54.0 & 58.6 & 82.5 \\
           \hspace{7pt}+ ToMe
                & 71.7 & \uline{68.8} & 60.7 & 47.9 & 67.4
                & 67.1 & \textbf{67.9} & 70.0 & 32.3 & \uline{62.8} & \uline{51.9} & 53.4
                & \uline{55.4} & 58.7 & 82.6 \\
           \hspace{7pt}+ MemForest
                & \uline{77.4} & 64.9 & \textbf{62.6} & 56.2 & \uline{70.7}
                & 62.9 & \uline{66.1} & 73.3 & 30.8 & 50.0 & 50.4 & 50.0
                & \textbf{55.5} & \uline{58.7} & \uline{82.6} \\
           \rowcolor{memyellow}
           \hspace{7pt}+ \textbf{RecordFolder (Ours)}
                & \textbf{77.5} & \textbf{69.1} & 60.1 & \textbf{60.4} & \textbf{71.3}
                & \uline{74.3} & 62.5 & \textbf{93.3} & \uline{33.8} & \textbf{65.4} & 45.9 & \textbf{54.4}
                & 55.2 & \textbf{60.3} & \textbf{84.8} \\

           \bottomrule
           \end{tabular}
           }
           \vspace{-2mm}
       \end{table*}

\begin{table*}[t]
    \caption{
    \textbf{Record compression on M3-Bench across compression ratios.}
    }
    \vspace{-2mm}
    \label{tab:main_m3agent}
    \centering
    \setlength{\tabcolsep}{3pt}
    \renewcommand{\arraystretch}{1.05}
    \resizebox{\textwidth}{!}{
    \begin{tabular}{lcccccc|cccccc|c|c}
    \toprule
    \multirow{2}{*}[-0.8ex]{\textbf{Method}}
         & \multicolumn{6}{c|}{\textbf{M3-Bench-robot}}
         & \multicolumn{6}{c|}{\textbf{M3-Bench-web}}
         & \multirow{2}{*}[-0.8ex]{\textbf{Avg.}}
         & \multirow{2}{*}[-0.8ex]{\textbf{\%}} \\
    \cmidrule(lr){2-7}\cmidrule(lr){8-13}
         & ME & MH & CM & PU & GK & \textbf{All}
         & ME & MH & CM & PU & GK & \textbf{All}
         &  &  \\
    \midrule

    \ub{M3-Agent}
         & \ub{30.9} & \ub{29.4} & \ub{29.6} & \ub{41.1} & \ub{22.6} & \ub{30.3}
         & \ub{44.9} & \ub{25.6} & \ub{44.8} & \ub{58.6} & \ub{53.7} & \ub{47.9}
         & \ub{39.1} & \ub{100.0} \\



    \midrule
    \multicolumn{15}{l}{\textit{Record compression ratio $\downarrow 50\%$}} \\

    \hspace{7pt}+ Random Pruning
         & 28.4 & 25.9 & 27.3 & 38.7 & 19.6 & 28.1
         & 35.6 & 23.0 & 34.9 & 52.3 & 44.2 & 41.2
         & 34.7 & 88.7 \\

    \hspace{7pt}+ KMeans
         & 29.1 & 30.6 & 28.4 & 41.4 & 20.8 & 29.2
         & 41.1 & 21.7 & 42.5 & 55.0 & 47.4 & 43.9
         & 36.6 & 93.6 \\

    \hspace{7pt}+ DART
         & 28.9 & 25.9 & 29.4 & 39.1 & \textbf{22.0} & 29.1
         & 39.9 & 23.6 & 39.4 & 52.9 & 47.2 & 43.1
         & 36.1 & 92.3 \\

    \hspace{7pt}+ StreamMeCo
         & \underline{32.3} & 28.2 & 29.8 & 41.4 & 21.1 & 30.6
         & 39.7 & \underline{25.2} & 38.9 & \underline{58.4} & 47.3 & 44.7
         & 37.7 & 96.2 \\

    \hspace{7pt}+ Random Merging
         & \textbf{33.0} & \underline{32.9} & \textbf{31.9} & 44.3 & 19.6 & 30.8
         & 41.8 & 21.0 & 40.1 & 58.0 & 46.3 & 44.6
         & 37.7 & 96.4 \\

    \hspace{7pt}+ ToMe
         & 30.5 & \underline{32.9} & 29.6 & 42.3 & 19.9 & 30.9
         & \underline{42.7} & 24.7 & 41.7 & 54.3 & \textbf{53.1} & 46.1
         & 38.5 & 98.5 \\

    \hspace{7pt}+ MemForest
         & 31.6 & 31.8 & \underline{31.1} & \underline{45.1} & \underline{21.4} & \textbf{31.6}
         & 42.6 & 23.9 & \underline{43.4} & 57.1 & 51.5 & \underline{46.5}
         & \underline{39.0} & \underline{99.7} \\

    \rowcolor{memyellow}
    \hspace{7pt}+ \textbf{RecordFolder (Ours)}
         & 31.7 & \textbf{34.1} & 30.0 & \textbf{45.3} & 19.9 & \underline{31.0}
         & \textbf{44.6} & \textbf{27.8} & \textbf{44.6} & \textbf{59.6} & \underline{52.4} & \textbf{48.2}
         & \textbf{39.6} & \textbf{101.3} \\

    \midrule
    \multicolumn{15}{l}{\textit{Record compression ratio $\downarrow 70\%$}} \\

    \hspace{7pt}+ Random Pruning
         & 27.3 & 25.9 & 26.7 & 39.2 & \textbf{20.8} & 27.7
         & 35.0 & 19.9 & 34.9 & 48.7 & 39.5 & 37.3
         & 32.5 & 83.1 \\

    \hspace{7pt}+ KMeans
         & 28.7 & 28.2 & 27.3 & 42.7 & 19.3 & 28.9
         & 35.5 & 20.1 & 34.9 & 39.7 & 42.5 & 39.1
         & 34.0 & 87.0 \\

    \hspace{7pt}+ DART
         & 27.8 & 25.9 & 25.4 & 40.0 & 16.2 & 27.2
         & 35.4 & 19.5 & 36.6 & 48.2 & 45.0 & 39.2
         & 33.2 & 84.9 \\

    \hspace{7pt}+ StreamMeCo
         & 29.1 & 30.6 & 26.9 & 40.3 & 19.9 & 28.4
         & 37.0 & 21.7 & 36.1 & 53.0 & 46.3 & 41.9
         & 35.1 & 89.8 \\

    \hspace{7pt}+ Random Merging
         & 30.6 & \underline{31.8} & 28.8 & \underline{45.1} & 19.0 & 30.4
         & 41.1 & 21.9 & 39.6 & \underline{58.6} & 41.5 & 43.2
         & 36.8 & 94.1 \\

    \hspace{7pt}+ ToMe
         & \underline{31.7} & 27.1 & 28.6 & 43.1 & \underline{20.2} & 30.6
         & 42.9 & 23.6 & \underline{43.6} & 56.5 & 49.8 & 45.7
         & 38.2 & 97.7 \\

    \hspace{7pt}+ MemForest
         & 31.2 & \textbf{32.9} & \textbf{32.4} & 43.6 & 17.4 & \underline{30.7}
         & \textbf{44.8} & \underline{23.9} & 38.9 & 57.3 & \underline{50.4} & \underline{46.3}
         & \underline{38.5} & \underline{98.5} \\

    \rowcolor{memyellow}
    \hspace{7pt}+ \textbf{RecordFolder (Ours)}
         & \textbf{32.3} & 29.4 & \underline{30.9} & \textbf{45.4} & 19.9 & \textbf{31.4}
         & \underline{44.5} & \textbf{28.9} & \textbf{44.1} & \textbf{60.3} & \textbf{51.6} & \textbf{48.7}
         & \textbf{40.1} & \textbf{102.6} \\

    \bottomrule
    \end{tabular}
    }
    \vspace{-4mm}
\end{table*}

\noindent \textbf{Results on Additional Memory Frameworks.}
Tables~\ref{tab:main_memos} and~\ref{tab:main_m3agent} evaluate RecordFolder on MemOS and M3-Agent.
On MemOS, it retains 91.0\% and 84.8\% of original performance with 50\% and 70\% fewer records.
On M3-Agent, mean accuracy across the robot and web subsets rises from 39.1 without compression to 39.6 and 40.1 at these reduction ratios.
This corresponds to 102.6\% performance retention at 70\% reduction.
Figure~\ref{fig:context_results}(c) evaluates ContextFolder on M3-Bench-robot at 50\% record-count reduction.
Across the three record compressors, BabelTele and LLMLingua-2 use 29.3--31.7 tokens per record, compared with 20.7--21.5 for ContextFolder.
After RecordFolder, ContextFolder matches LLMLingua-2 at 29.47\% accuracy, using 20.9 versus 31.7 tokens per record, a reduction of 34.1\%.
BabelTele achieves higher accuracy at 32.13\% but uses 30.2 tokens per record.
Short, information-dense event records may leave little redundant text for token deletion or symbolic rewriting.
ContextFolder compresses shared descriptions across grouped records, with fact-recovery checks guiding revision.
RecordFolder reduces the number of textual retrieval units, while ContextFolder shortens their contents within M3-Agent's multimodal memory graph.

\subsection{Ablation Studies and Analysis}
\label{subsec:ablation}

\noindent \textbf{Effects of Each Design.}
Table~\ref{tab:ablation_anchor} evaluates anchor selection in RecordFolder.
DPC achieves the best average accuracy of 57.2 and the highest LoCoMo score of 59.3, while MinMax leads on LongMemEval with 55.6.
This indicates that progressive density-and-novelty selection balances local representativeness and complementary coverage.
Table~\ref{tab:ablation_affinity} isolates the assignment signals.
The full model averages 57.2; removing cosine similarity causes the largest drop of 3.4 points, compared with 1.9, 1.1, and 0.6 points for gain gating, entity affinity, and temporal chaining.
These results identify embedding similarity as the main semantic signal, with the other terms refining entity and temporal consistency.
Table~\ref{tab:ablation_action} compares pruning and merging with DPC anchors fixed.
Merging averages 57.2, 6.1 points above pruning and only 0.9 and 0.2 points below the uncompressed scores on LoCoMo and LongMemEval.
This mirrors Tables~\ref{tab:main_mem0} and~\ref{tab:main_memos}, where merging also performs better under record reduction.
Table~\ref{tab:ablation_context} evaluates ContextFolder after 50\% record reduction.
The full model achieves 56.6 and 51.8 accuracy with 12.31 and 13.40 tokens per record on LoCoMo and LongMemEval.
Removing the planner lowers accuracy by 0.7 and 0.8 points while increasing token use; removing the verifier saves tokens but costs 1.9 and 1.4 points.
The planner thus improves accuracy and compression, while the verifier strengthens recovery at a small token cost.
Overall, RecordFolder forms stronger retrieval units, and ContextFolder further compresses their contents.

\begin{table}[t]
\centering
\setlength{\tabcolsep}{3.5pt}
\renewcommand{\arraystretch}{1.10}
\captionsetup{font=small,labelfont=bf,skip=2pt}

\begin{minipage}[t]{0.49\textwidth}
\vspace{0pt}
\centering
\captionof{table}{\textbf{Anchor selection ablation.} Accuracy at 50\% record compression.}
\label{tab:ablation_anchor}
\footnotesize
\renewcommand{\arraystretch}{1.32}
\begin{tabularx}{\linewidth}{@{}l
>{\hsize=0.9\hsize\centering\arraybackslash}X
>{\hsize=1.35\hsize\centering\arraybackslash}X
>{\hsize=0.75\hsize\centering\arraybackslash}X@{}}
\toprule
\textbf{Settings} & \textbf{LoCoMo} & \textbf{LongMemEval} & \textbf{Average} \\
\midrule
Random & 56.7 & 53.0 & 54.8 \\
KMeans & 56.5 & 54.4 & 55.4 \\
MinMax & 58.0 & \textbf{55.6} & 56.8 \\
\rowcolor{memyellow}
\textbf{DPC} & \textbf{59.3} & 55.0 & \textbf{57.2} \\
\bottomrule
\end{tabularx}
\end{minipage}
\hfill
\begin{minipage}[t]{0.49\textwidth}
\vspace{0pt}
\centering
\captionof{table}{\textbf{Affinity and chaining ablation.} Accuracy at 50\% record-count reduction.}
\label{tab:ablation_affinity}
\footnotesize
\begin{tabularx}{\linewidth}{@{}l
>{\hsize=0.9\hsize\centering\arraybackslash}X
>{\hsize=1.35\hsize\centering\arraybackslash}X
>{\hsize=0.75\hsize\centering\arraybackslash}X@{}}
\toprule
\textbf{Settings} & \textbf{LoCoMo} & \textbf{LongMemEval} & \textbf{Average} \\
\midrule
w/o $s_{\mathrm{ent}}$ & 58.4 & 53.8 & 56.1 \\
w/o $\cos$ & 55.6 & 52.0 & 53.8 \\
w/o gain gate & 57.6 & 53.0 & 55.3 \\
w/o chaining & 59.1 & 54.2 & 56.6 \\
\rowcolor{memyellow}
\textbf{Ours} & \textbf{59.3} & \textbf{55.0} & \textbf{57.2} \\
\bottomrule
\end{tabularx}
\end{minipage}

\vspace{3pt}

\begin{minipage}[t]{0.49\textwidth}
\vspace{0pt}
\centering
\captionof{table}{\textbf{Pruning versus merging.} Accuracy at 50\% reduction with fixed DPC anchors.}
\label{tab:ablation_action}
\footnotesize
\renewcommand{\arraystretch}{1.10}
\begin{tabularx}{\linewidth}{@{}l
>{\hsize=0.9\hsize\centering\arraybackslash}X
>{\hsize=1.35\hsize\centering\arraybackslash}X
>{\hsize=0.75\hsize\centering\arraybackslash}X@{}}
\toprule
\textbf{Settings} & \textbf{LoCoMo} & \textbf{LongMemEval} & \textbf{Average} \\
\midrule
Vanilla & 60.2 & 55.2 & 57.7 \\
Pruning  & 54.7 & 47.6 & 51.1 \\
\rowcolor{memyellow}
\textbf{Merging} & \textbf{59.3} & \textbf{55.0} & \textbf{57.2} \\
\bottomrule
\end{tabularx}
\end{minipage}
\hfill
\begin{minipage}[t]{0.49\textwidth}
\vspace{0pt}
\centering
\captionof{table}{\textbf{ContextFolder ablation.} Accuracy (\%) / tokens per record after 50\% record   reduction.}
\label{tab:ablation_context}
\footnotesize
\renewcommand{\arraystretch}{1.10}
\begin{tabularx}{\linewidth}{@{}l*{2}{>{\centering\arraybackslash}X}@{}}
\toprule
\textbf{Settings} & \textbf{LoCoMo} & \textbf{LongMemEval} \\
\midrule
w/o planner & 55.9 / 16.25 & 51.0 / 14.44 \\
w/o verifier & 54.7 / 11.94 & 50.4 / 13.19 \\
\rowcolor{memyellow}
\textbf{Ours} & \textbf{56.6} / 12.31 & \textbf{51.8} / 13.40 \\
\bottomrule
\end{tabularx}
\end{minipage}
\vspace{-2mm}
\end{table}

\begin{figure}[!t]
    \centering
    \includegraphics[width=\linewidth]{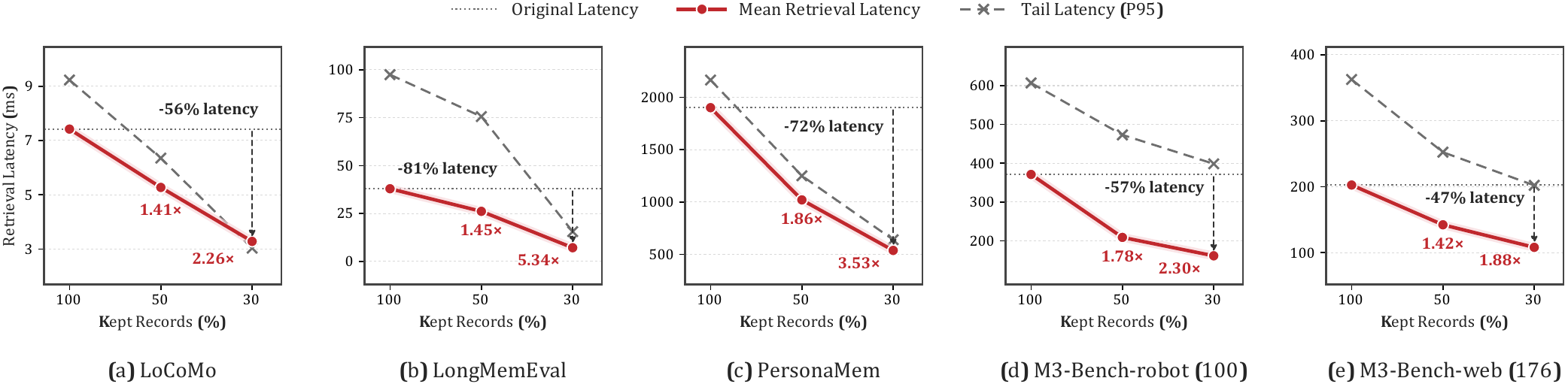}
    \vspace{-5mm}
    \caption{\textbf{Retrieval efficiency under record compression.}
    Mean and 95th-percentile latency with 100\%, 50\%, and 30\% of records retained.
    Dotted lines indicate uncompressed mean latency.
    Keeping 30\% yields 1.88--5.34$\times$ mean speedups across five settings.}
    \label{fig:efficiency}
    \vspace{-2mm}
\end{figure}

\noindent \textbf{Efficiency Analysis.}
Figure~\ref{fig:efficiency} reports retrieval latency across three text-based and two multimodal settings.
Mean speedups rise from $1.41$--$1.86\times$ at 50\% retention to \textbf{1.88--5.34$\times$} at 30\%.
At 30\% retention, mean latency drops by 56\%, 81\%, and 72\% on LoCoMo, LongMemEval, and PersonaMem, respectively; PersonaMem falls from 1,900 to 538\,ms.
The largest gain appears on LongMemEval, where the mean speedup grows from $1.45\times$ at 50\% retention to $5.34\times$ at 30\%, while tail latency falls from 97.5 to 15.5\,ms.
This parallel reduction in mean and tail latency shows that record compression alleviates expensive retrieval cases, not only the average query cost.
M3-Bench-robot and M3-Bench-web reach $2.30\times$ and $1.88\times$ speedups, respectively, although their tail-latency gains are smaller than those on the text benchmarks.
This gap suggests that record retrieval is only one part of the latency in multimodal settings, where additional processing costs remain.
Record compression therefore reduces both average and high-latency retrieval overhead as the external memory grows.


\section{Conclusion}
\vspace{-5pt}
Growing external memory stores increase the storage, retrieval, and generation costs of long-term agents.
To address these costs, we propose MemFolder, a structure-guided memory folding framework comprising two components: RecordFolder and ContextFolder.
The former reduces record count by grouping and consolidating related evidence using semantic, entity, and temporal relations.
The latter further reuses these groups for compact encoding, with fact-recovery checks guiding revisions.
Experiments across multiple benchmarks demonstrate the strong performance of RecordFolder under record compression.
Combined with ContextFolder, existing record compression methods further shorten retrieval units, reducing both record count and context length.
\subsection*{AI use statement}

Generative AI tools were used to assist with manuscript drafting, language polishing, and structural editing, as well as with literature search and code development.

\bibliography{iclr2027_conference}
\bibliographystyle{iclr2027_conference}

\appendix
\clearpage
\begingroup
\renewcommand{\topfraction}{0.95}
\renewcommand{\bottomfraction}{0.90}
\renewcommand{\textfraction}{0.05}
\renewcommand{\floatpagefraction}{0.75}
\section*{Appendix}
\pdfbookmark[0]{Appendix}{memfolder-appendix}
\makeatletter
\let\ocSavedAddtocontents\addtocontents
\renewcommand{\addtocontents}[2]{%
  \def\ocCurrentExtension{#1}\def\ocMainContents{toc}%
  \ifx\ocCurrentExtension\ocMainContents
    \ocSavedAddtocontents{apx}{#2}%
  \else
    \ocSavedAddtocontents{#1}{#2}%
  \fi
}
\setcounter{tocdepth}{2}
\begingroup
\hypersetup{linkcolor=black}
\color{black}
\@starttoc{apx}
\endgroup
\makeatother
\medskip

\section{Additional Experimental Details and Analysis}
\label{app:additional-details}

\paragraph{Memory Framework Details.}
We consider both text-based and multimodal memory frameworks:
\begin{itemize}
\item \textbf{Mem0}~\citep{chhikara2025mem0} is a long-term memory system that extracts, updates, stores, and retrieves persistent textual records from agent interactions. We use Mem0 as the primary text-based framework and construct an uncompressed memory store before applying each compression method.

\item \textbf{MemOS}~\citep{li2025memos} is a memory operating system for AI systems that organizes multiple forms of memory, including plaintext, activation, and parametric memory, under a unified management layer. It provides a framework-level setting for studying record compression beyond a single text-memory implementation.

\item \textbf{M3-Agent}~\citep{long2026m3-agent} is a multimodal agent framework that continuously processes video and audio inputs and builds an entity-centric long-term memory. Its memory integrates visual, auditory, and textual information, which supports multi-turn retrieval and reasoning over dynamic environments.
\end{itemize}

\paragraph{Benchmark Details.}
We evaluate long-term memory compression on three conversational benchmarks and two multimodal video benchmarks:
\begin{itemize}
\item \textbf{LoCoMo}~\citep{maharana2024locomo} contains long multi-session conversations and questions covering single-hop, multi-hop, temporal, and open-domain reasoning, testing whether an agent can recover evidence distributed across an extended interaction history.

\item \textbf{LongMemEval}~\citep{wu2025longmemeval} evaluates long-term interactive memory with single-session and multi-session questions, preference tracking, knowledge updates, and temporal reasoning.

\item \textbf{PersonaMem}~\citep{jiang2025personamem} benchmarks dynamic user profiling and personalized response generation. We use its 32K setting, which tests user-fact recall, preference evolution, recommendation, and generalization across scenarios.

\item \textbf{M3-Bench-robot}~\citep{long2026m3-agent} is an online first-person video benchmark collected from daily robot environments. Its questions require multi-evidence, multi-hop, cross-modal, person-centered, and general-knowledge reasoning over continuously observed interactions.

\item \textbf{M3-Bench-web}~\citep{long2026m3-agent} is an offline web-video benchmark spanning diverse video categories. It evaluates the same long-term multimodal reasoning abilities over prerecorded videos with questions that require evidence from multiple moments and modalities.
\end{itemize}

\paragraph{Baseline Details.}
We compare record compression against pruning and merging baselines:
\begin{itemize}
\item \textbf{Random Pruning} randomly discards records until the target keep ratio is reached. It isolates the effect of record removal without using content, temporal, or graph information.

\item \textbf{KMeans}~\citep{kmeans} clusters record embeddings into the target number of groups and retains the record closest to each cluster centroid, providing an embedding-space representative-selection baseline.

\item \textbf{DART}~\citep{wen2025dart} is a diversity-based pruning method that removes highly redundant representations by comparing them with a small set of pivot representations. We adapt this operation from visual tokens to memory records.

\item \textbf{StreamMeCo}~\citep{wang2026streammeco} compresses structured memory graphs with a dual strategy: it samples representatives from isolated nodes and prunes connected nodes using entity importance and semantic similarity.

\item \textbf{Random Merging} partitions records into retained and removed sets, then randomly assigns removed records to retained records for consolidation until the target keep ratio is reached.

\item \textbf{ToMe}~\citep{bolya2023tome} progressively matches similar representations across two partitions and merges each matched pair. We adapt this token merging procedure to records by matching each record with its most similar counterpart.

\item \textbf{MemForest}~\citep{wang2026memforest} combines global semantic similarity with local temporal continuity to partition records into event-centric units. It constructs a maximum-spanning-tree EventTree for each unit and progressively merges high-weight pairs, while using anchor-guided propagation to retrieve relevant temporal neighbors.
\end{itemize}

\subsection{Implementation Details}\label{app:implementation}

All compression methods start from the same uncompressed memory store built with each memory framework. For Mem0~\citep{chhikara2025mem0} and MemOS~\citep{li2025memos}, we use \texttt{gpt-4o-mini} for memory extraction and updating, and
     \texttt{text-embedding-3-small} for record embeddings, following the original
     papers. For M3Agent~\citep{long2026m3-agent}, we deploy \texttt{gpt-4o} for evaluation and use
     \texttt{text-embedding-3-large} for record embeddings, as in the original
     paper. The resulting snapshots are shared by all methods before compression.
     For each target reduction ratio, methods operate on the same records and are
     evaluated with the same retrieval and question-answering procedure. We
     preserve the original question sets and report accuracy together with
     record-count and token-level statistics. Answer quality on Mem0 and MemOS is
     judged with \texttt{gpt-4o-mini}, whereas M3Agent is evaluated with
     \texttt{gpt-4o}. All methods therefore share a common memory and evaluation
     protocol within each framework.

\noindent \textbf{Compression Settings.}
RecordFolder uses \texttt{gpt-4o-mini} for entity grounding, an entity-label similarity threshold $\theta_{\mathrm{ent}}=0.85$, and a neighborhood cap $k_{\mathrm{nn}}=5$.
Temporal refinement uses a predecessor window $W=50$ and at most $d=8$ positive-gain links per successor; both signal variances are floored at $10^{-6}$.
ContextFolder assigns weights of $3$, $2$, and $1$ to high-, medium-, and low-importance facts and halves the weights of inferable facts.
Its medium-importance requirement threshold is $\tau_{\mathrm{req}}=0.65$, its minimum token budget is $3$, and its budget tolerance is $\epsilon_B=0.15$.
The default iteration limit is $R_{\max}=3$, except when varied in the iteration analysis.

\subsection{Hyperparameter Analysis}
\label{subsec:hyperparameter_analysis}

We study the sensitivity of RecordFolder's affinity weighting and ContextFolder's iteration limit on LoCoMo and LongMemEval. Figure~\ref{fig:hyperparameter_analysis} summarizes the results.

\begin{figure*}[h]
\centering
\includegraphics[width=\textwidth]{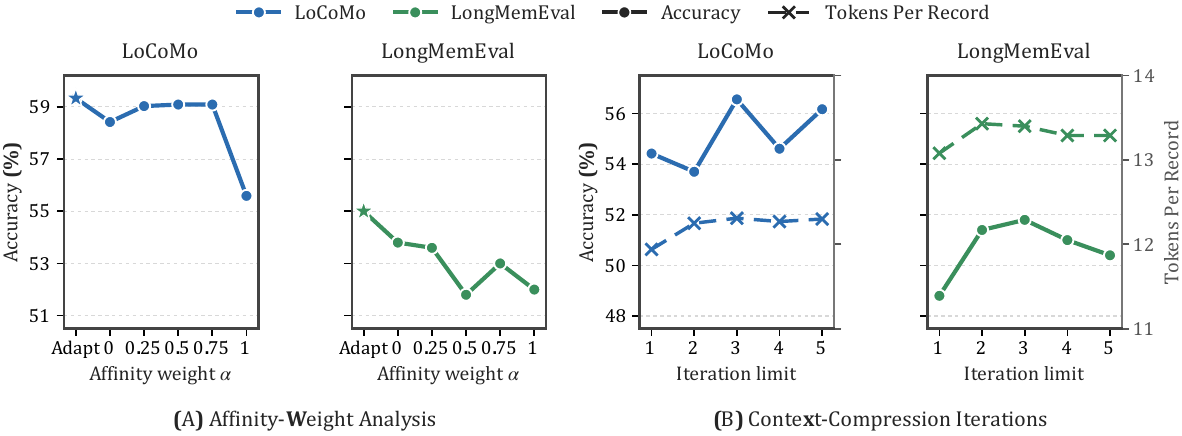}
\vspace{-5mm}
\caption{\textbf{Hyperparameter analysis on LoCoMo and LongMemEval.}
(A) Affinity-weight analysis (panels 1--2); (B) ContextFolder iteration analysis (panels 3--4), each ordered by LoCoMo then LongMemEval.
Blue and green denote the benchmarks; solid circles show accuracy and dashed crosses show tokens per record.
Higher accuracy and lower token usage are preferred.}
\vspace{-2mm}
\label{fig:hyperparameter_analysis}
\end{figure*}

For affinity weighting, the self-adaptive scheme achieves 59.33\% on LoCoMo and 55.0\% on LongMemEval. The best fixed settings reach 59.09\% on LoCoMo ($\alpha=0.5$ or $0.75$) and 53.8\% on LongMemEval ($\alpha=0$), showing a modest gain on LoCoMo and a clearer gain on LongMemEval.

For ContextFolder, three iterations yield the highest accuracy on both benchmarks: 56.56\% on LoCoMo and 51.8\% on LongMemEval, up from 54.42\% and 48.8\% with one iteration. At three iterations, token usage is 12.31 and 13.40 tokens per record; across the sweep, it varies only from 11.94 to 12.31 and from 13.08 to 13.43, respectively. More iterations do not improve accuracy or materially reduce token use, so we use three iterations by default.

\section{ContextFolder Compression Traces}
\label{app:traces}

This appendix records three groups compressed end to end on LoCoMo, taken from the
production run and reproduced without modification. Each trace lists the source
records, the facts and questions the planner produced from them, every encoding the
compressor wrote, the blind reader's answers and the verifier's verdict on each round,
and the encoding finally stored. Token counts are those recorded by the implementation.
All three traces use $\tau=0.8$. Their budgets are $B_G=\max\{3,\operatorname{round}(\tau S_G)\}$,
where $S_G=\sum_{m\in G}\ell(m)$. A round is accepted when all required checks pass and the encoding fits within
$(1+\epsilon_B)B_G$, with $\epsilon_B=0.15$.

\subsection{Case A: Acceptance after Three Rounds}
\label{app:case-a}

This group contains four records totaling $48$ tokens at $\kappa=0.7$, with a token budget of $38$.
The four records say nearly the same thing four times, so the difficulty is not length
but deciding which of the overlapping phrasings still have to be recoverable.

\begin{table}[h]
\centering
\small
\begin{tabular}{clr}
\toprule
\# & Record & Tokens \\
\midrule
1 & User appreciates having people around who accept and support them & 11 \\
2 & User appreciates their support network for love, guidance, and acceptance & 13 \\
3 & User appreciates support and is grateful for sharing their journey and helping others & 14 \\
4 & User appreciates the support of those close to them & 10 \\
\bottomrule
\end{tabular}
\end{table}

The planner marked all four facts \textit{high} and \textit{required}, pairing each with
a question; fact 4 carried \textit{Who does the user appreciate support from?}, expecting
\textit{those close to them}.

\paragraph{Round 1 (19 tokens).}

{\footnotesize
\begin{verbatim}
E: appreciate.support.network
E: appreciate.support.close
E: grateful.sharing.journey
\end{verbatim}
}

\noindent The blind reader recovered facts 1--3 and answered \textsc{unknown} for fact 4,
so the round was returned with one must-fix item:

{\footnotesize
\begin{verbatim}
MUST-FIX (required facts the blind reader could not recover):
  * User appreciates the support of those close to them.
    Express THESE explicitly; clarity first for them.
    Re-encode by EDITING the previous encoding. Restore ONLY the facts
    listed as MUST-FIX; do not spend tokens on optional facts unless you
    are well under budget; if over budget, tighten wording or drop
    optional facts.
\end{verbatim}
}

\paragraph{Round 2 (26 tokens).} The compressor edited its previous encoding in place,
adding \texttt{E: appreciate.support.acceptance}. Fact 4 was still answered
\textsc{unknown}, and the same item was returned a second time.

\paragraph{Round 3 (21 tokens).}

{\footnotesize
\begin{verbatim}
E: appreciate.support.acceptance
E: appreciate.support.love.guidance
E: appreciate.support.close
\end{verbatim}
}

\noindent The verifier marked all four checks as passed, giving a weighted recovery score of $1.0$.
The 21-token encoding also satisfied $21\leq1.15\times38$, and the loop terminated after the third round.

\paragraph{Stored.} Round 3 is stored at $21$ tokens, a retention ratio of $0.44$.
Round 1 is two tokens shorter but fails a required check.
The stored encoding is the shortest candidate that passes all required checks in this example.

\subsection{Case B: One Round on a Redundant Group}
\label{app:case-b}

This group contains six records totaling $62$ tokens at $\kappa=0.7$, with a token budget of $50$.
The six records contain overlapping descriptions of three core facts, which the planner consolidates before encoding.

\begin{table}[h]
\centering
\small
\begin{tabular}{cl}
\toprule
\# & Record \\
\midrule
1 & Volunteering is an important part of Caroline's life \\
2 & Caroline is carrying on volunteering at the youth center \\
3 & Volunteering at the youth center is an important part of Caroline's life \\
4 & Caroline wants to keep making a difference \\
5 & Caroline is carrying on volunteering at the youth center \\
6 & Caroline wants to keep making a difference \\
\bottomrule
\end{tabular}
\end{table}

The planner returned three facts: volunteering matters to her \textit{(high)}, she
volunteers at the youth center \textit{(high)}, and she wants to keep making a difference
\textit{(med)}.

\paragraph{Round 1 (17 tokens), accepted.}

{\footnotesize
\begin{verbatim}
E: volunteering important
A -> volunteering at youth center
? keep making a difference
\end{verbatim}
}

\noindent The verifier accepted all three checks in the first round, and the stored encoding retained $0.27$ of the source tokens.

\subsection{Case C: A Substituted Answer Caught by the Judge}
\label{app:case-c}

This group contains twelve records totaling $92$ tokens at $\kappa=0.3$, with a token budget of $74$.
This is the group cited in the ContextFolder section. Records 4 and 5, 6 and 7, and
9 and 10 are near-duplicates, and timestamps are excluded from the fact list by rule, so
the planner reduced twelve records to six facts.

\begin{table}[h]
\centering
\small
\begin{tabular}{cll}
\toprule
\# & Record & Date \\
\midrule
1 & Cares about mental health & 27 Jun 2023 \\
2 & User is keen on counseling or working in mental health & 27 Jun 2023 \\
3 & Values counseling and support groups & 27 Jun 2023 \\
4 & Thinking about taking care of mental health & 25 May 2023 \\
5 & Values taking care of mental health & 25 May 2023 \\
6 & Interested in counseling and mental health & 3 Jul 2023 \\
7 & Keen on counseling or working in mental health & 3 Jul 2023 \\
8 & Looking into counseling and mental health jobs & 12 Jul 2023 \\
9 & Looking into counseling and mental health as a career & 12 Jul 2023 \\
10 & Looking into counseling and mental health as a career & 12 Jul 2023 \\
11 & Started looking into counseling and mental health career options to help others & 12 Jul 2023 \\
12 & Values sharing stories and giving others hope & 9 Jun 2023 \\
\bottomrule
\end{tabular}
\end{table}

\begin{table}[h]
\centering
\footnotesize
\begin{tabular}{cp{0.42\textwidth}p{0.38\textwidth}}
\toprule
\# & Fact (level) & Question $\rightarrow$ expected answer \\
\midrule
1 & Cares about mental health \textit{(high)} & What does the user care about? $\rightarrow$ Mental health \\
2 & Keen on counseling or working in mental health \textit{(high)} & Interest regarding career? $\rightarrow$ Counseling or mental health work \\
3 & Values counseling and support groups \textit{(med)} & What does the user value? $\rightarrow$ Counseling and support groups \\
4 & Looking into counseling and mental health jobs \textit{(high)} & Currently exploring? $\rightarrow$ Counseling and mental health jobs \\
5 & Started looking into these options to help others \textit{(med)} & Why? $\rightarrow$ To help others \\
6 & Values sharing stories and giving others hope \textit{(med)} & Values in support? $\rightarrow$ Sharing stories, giving hope \\
\bottomrule
\end{tabular}
\end{table}

\paragraph{Round 1 (44 tokens).} The compressor came in well under budget by writing five
lines and omitting fact 3.

{\scriptsize
\begin{verbatim}
cares about mental health
keen on counseling or working in mental health
looking into counseling and mental health jobs
started looking into counseling and mental health career options to help others
values sharing stories and giving others hope
\end{verbatim}
}

\noindent Questions 1, 2, 4 and 5 were answered correctly. Question 3 asked what the user
valued in relation to mental health; the encoding no longer carried fact 3, and the blind
reader answered \textit{sharing stories and giving others hope} --- a fact the encoding
did carry, offered in place of the one that was lost. Nothing in that answer looks like a
failure on its own, and a check that only watched for \textsc{unknown} would have passed
it; the judging call compared it against the planner's expected answer and marked it
must-fix. Question 6 was answered \textsc{unknown}. Both items were returned to the
compressor:

{\footnotesize
\begin{verbatim}
MUST-FIX (required facts the blind reader could not recover):
  * User values counseling and support groups.
  * User values sharing stories and giving others hope.
    Express THESE explicitly; clarity first for them.
    Re-encode by EDITING the previous encoding. Restore ONLY the facts
    listed as MUST-FIX; do not spend tokens on optional facts unless you
    are well under budget; if over budget, tighten wording or drop
    optional facts.
\end{verbatim}
}

\paragraph{Round 2 (50 tokens), accepted.}

{\scriptsize
\begin{verbatim}
cares about mental health
keen on counseling or working in mental health
values counseling and support groups
looking into counseling and mental health jobs
started looking into counseling and mental health career options to help others
values sharing stories and giving others hope
\end{verbatim}
}

\noindent All six questions were recovered and the loop stopped early, at a ratio of
$0.54$.

\subsection{On the Reliability of the Check}
\label{app:check-noise}

The reader can fail to recover a fact that is present in the encoding.
In Case~C, fact 6 appears in both rounds, but the reader answers \textsc{unknown} in round 1 and passes the check in round 2.
Case~A shows a similar difference for \texttt{E: appreciate.support.close}, which appears in rounds 1 and 3.

Such recovery failures can trigger additional revisions.
An omitted fact that incorrectly passes verification can leave information loss undetected.
In Case~C, the judge flags question 3 for revision after comparing the response with the expected answer.
The blind reader receives no source records.

\clearpage
\section{Prompt Templates}
\label{app:prompts}

We provide the prompts used by the main LLM components in execution order.

\begin{figure*}[!htbp]
\centering
\begin{memprompt}{Entity Grounding Prompt: Initial Session}

Role: exhaustive entity extraction for a memory compression graph.

You will ground every memory node of one conversation session. The extracted
entities decide which memories are allowed to merge, so recall of
content-bearing entities matters more than precision of wording.

Session metadata:
- sample_id: {sample_id}
- exact_timestamp: {timestamp}

Extraction rules:
1. Extract ALL content-bearing entities mentioned or clearly implied by the
   node text: people (including authors/creators/ public figures), named works
   (books, movies, songs, shows) together with their creators, places and
   locations, organizations / communities / groups, pets or named animals,
   significant objects / possessions, activities / hobbies / jobs /
   fields of work, topics or concepts someone could ask about later, and
   concrete events.
2. The bare word "User" refers to the source speaker; resolve it to the
   speaker's own name and do not emit "user" as an entity.
3. Do NOT stop at the protagonist. A node like "Loved 'X' by Y" must yield
   both the work X and the person Y. Aim for at least 2 entities per node
   whenever the text carries any concrete content; only pure greetings or
   chit-chat may have fewer.
4. `canonical` must be a short noun phrase in lowercase, without a leading
   "the", without verbs. Aliases of the same thing share one canonical label.
5. Do not invent entities that are not in the text.
6. `relations` are optional; keep them only when clean, with a noun-phrase
   subject and object (never a verb phrase or sentence).
7. Set `opens_channel=true` for any entity that could plausibly recur in
   later memories.

Allowed entity types: {types}

Return exactly one JSON object:
{
  "nodes": [
    {
      "id": 0,
      "entities": [
        {"canonical": "becoming nicole", "type": "work", "roles": ["liked"], "mentions": ["Becoming Nicole"], "opens_channel": true},
        {"canonical": "amy ellis nutt", "type": "person", "roles": ["author"], "mentions": ["Amy Ellis Nutt"], "opens_channel": true}
      ],
      "relations": [{"subject": "amy ellis nutt", "predicate": "wrote", "object": "becoming nicole"}]
    }
  ]
}

Every input id must occur exactly once. Output JSON only, no commentary.

INPUT NODES:
{records}

\end{memprompt}
\caption{\textbf{Initial-session entity grounding prompt.} The prompt extracts
content-bearing entities from one conversation session without an earlier
canonical-label registry.}
\label{fig:prompt_grounding_initial}
\end{figure*}

\begin{figure*}[!htbp]
\centering
\begin{memprompt}{Entity Grounding Prompt: Cross-Session Registry}

Role: exhaustive entity extraction for a memory compression graph.

You will ground every memory node of one conversation session. The extracted
entities decide which memories are allowed to merge, so recall of
content-bearing entities matters more than precision of wording.

Session metadata:
- sample_id: {sample_id}
- exact_timestamp: {timestamp}
- known/recurrent canonical labels from earlier sessions: {registry}

Extraction rules:
1. Extract ALL content-bearing entities mentioned or clearly implied by the
   node text: people (including authors/creators/public figures), named works
   (books, movies, songs, shows) together with their creators, places and
   locations, organizations / communities / groups, pets or named animals,
   significant objects / possessions, activities / hobbies / jobs /
   fields of work, topics or concepts someone could ask about later, and
   concrete events.
2. The bare word "User" refers to the source speaker; resolve it to the
   speaker's own name and do not emit "user" as an entity.
3. Do NOT stop at the protagonist. A node like "Loved 'X' by Y" must yield
   both the work X and the person Y. Aim for at least 2 entities per node
   whenever the text carries any concrete content; only pure greetings or
   chit-chat may have fewer.
4. `canonical` must be a short noun phrase in lowercase, without a leading
   "the", without verbs. Aliases of the same thing share one canonical label.
5. Coreference discipline:
   a. If an entity is the SAME referent as a known/recurrent label above,
      reuse that label EXACTLY.
   b. If it merely shares words but is a DIFFERENT referent (e.g. a different
      person with the same first name, different kids, a different song),
      create a disambiguated label, e.g. "max (neighbor)" vs "max (coworker)".
   c. Generic referents must carry owner qualification when an owner is
      known: "melanie's kids", not bare "kids"; "caroline's garden", not
      "garden". Truly public/generic concepts (art, pottery, mental health)
      stay unqualified.
6. Do not invent entities that are not in the text.
7. `relations` are optional; keep them only when clean, with a noun-phrase
   subject and object (never a verb phrase or sentence).
8. Set `opens_channel=true` for any entity that could plausibly recur in
   later memories.

Allowed entity types: {types}

Return exactly one JSON object:
{
  "nodes": [
    {
      "id": 0,
      "entities": [
        {"canonical": "becoming nicole", "type": "work", "roles": ["liked"], "mentions": ["Becoming Nicole"], "opens_channel": true},
        {"canonical": "amy ellis nutt", "type": "person", "roles": ["author"], "mentions": ["Amy Ellis Nutt"], "opens_channel": true}
      ],
      "relations": [{"subject": "amy ellis nutt", "predicate": "wrote", "object": "becoming nicole"}]
    }
  ]
}

Every input id must occur exactly once. Output JSON only, no commentary.

INPUT NODES:
{records}

\end{memprompt}
\caption{\textbf{Cross-session entity grounding prompt.} The prompt adds
canonical-label reuse and owner-qualified coreference resolution.}
\label{fig:prompt_grounding_registry}
\end{figure*}

\begin{figure*}[!htbp]
\centering
\begin{memprompt}{Memory Merge Prompt}

You are an intelligent memory summarization assistant. Please extract the core information from the following {N} memory entries and their timestamps, and
generate a concise and coherent summary paragraph. If the memories contain relative time expressions (e.g., "last June," "two months ago"), convert them into
specific dates based on the corresponding timestamps (for example, if a memory mentions "last June" and the timestamp is February 2023, the actual time should be
June 2022). The summary length should be kept within the combined length of the original memories as much as possible. Output only the final summarized result,
without any explanations, analysis, reasoning steps, or additional content.
  Memory 1: {text}, Timestamp 1: {ts}
  ...
  Memory N: {text}, Timestamp N: {ts}

\end{memprompt}
\caption{\textbf{Memory merge prompt.} The LLM consolidates records assigned
to one RecordFolder group and resolves relative dates from timestamps.}
\label{fig:prompt_merge}
\end{figure*}

\begin{figure*}[!htbp]
\centering
\begin{memprompt}{ContextFolder Planner Prompt}

From the memory entries below, extract the KEY facts as a numbered list, and for each fact write a short verification question with its expected answer. A
reader should be able to answer the question IF AND ONLY IF the fact is recoverable from a compressed encoding.

Scale the fact count to the input size: about 1 atomic fact per 10 source tokens (min 2, max 12), one question per atomic fact. A short single memory deserves
only 2-4 questions about its ONE core fact (e.g. who/what/when), not a decomposition into trivia. For each fact also provide:
  - "importance": "high" for key events, final states, exact dates/numbers/names, entity bindings, negations; "med" for preferences, plans, relation directions,
    quantities; "low" for reasons, modifiers, emotional framing, superseded older states
  - "inferable": true if a reader could infer this fact from the OTHER extracted facts or from common knowledge, else false
Do NOT include meta information such as the memory entries' own timestamps as facts. Keep each fact atomic.

Memories:
{lines}

Answer with JSON only: {"facts": [{"fact": "...", "question": "...", "answer": "...", "importance": "high|med|low", "inferable": true|false}]}

\end{memprompt}
\caption{\textbf{Planner prompt.} The planner extracts atomic facts and creates
blind verification questions without access to downstream test queries.}
\label{fig:prompt_planner}
\end{figure*}

\begin{figure*}[!htbp]
\centering
\begin{memprompt}{ContextFolder Compressor Prompt}
You are a memory compressor writing for another equally intelligent LLM, not for humans. Compress the {n} memory entries below into a compact token sequence
that another LLM can decode WITHOUT any external codebook.

Fixed grammar (one fact per line):
- `@YYYY-MM-DD` ONLY when a fact itself contains time information; NEVER attach entry timestamps to timeless facts
- `A -> B` for state changes; `E: state` for attributes; `~` negation; `?` uncertain
- Exact entity full names; exact numbers/dates verbatim
- NEVER add `User:`/`She:` prefixes unless the source itself names the subject
- cl100k discipline: dotted fragments like `M.son.acc` are EXPENSIVE, never use them; common whole words are cheap
- Short single memories must come out strictly shorter than the source: keep verb + object, drop titles' authors, reasons, feelings. E.g. "Loved 'Becoming
Nicole' by Amy Ellis Nutt" -> `loved 'Becoming Nicole'`

SOFT TOKEN BUDGET: about {target} tokens (source: {src}).

TRIAGE THE FACTS BEFORE ENCODING. The key facts are listed below with labels:
- MUST KEEP: [high] facts that are NOT inferable — key events, final states, exact names/dates/numbers, negations. These must survive verbatim-recoverable.
- DROP FIRST: facts marked [inferable] — the reader can re-derive them from the facts you keep or from common knowledge. Do not spend a single token on them.
- DROP IF TIGHT: [low] facts — reasons, modifiers, emotional framing, superseded older states.
- [med] facts: keep only while you are within budget.
Omitting a fact is allowed and expected; distorting or inventing one is not.

Key facts to triage:
{facts_list}

Output ONLY the encoding.

Memories:
{lines}

\end{memprompt}
\caption{\textbf{Compressor prompt.} The compressor writes a compact encoding
under a soft token budget and prioritizes non-inferable facts.}
\label{fig:prompt_compressor}
\end{figure*}

\begin{figure*}[!htbp]
\centering
\begin{memprompt}{ContextFolder Verifier: Blind Answerer}

You see ONLY a compact memory encoding and a list of questions. The encoding uses this grammar: `@YYYY-MM-DD` = the fact happened on that date; `A -> B` = A
changed into / transferred to B; `E: x` = entity E has attribute/state x; `~` = negation; `?` = uncertain/rumored; `;` and newlines separate facts; filler words
are dropped but entity names, numbers, dates and relation directions are preserved. First silently decode each line into plain facts, then answer each question
using ONLY what can be inferred from the decoded facts. If a question cannot be answered from the encoding, answer exactly "UNKNOWN". Do not guess.

Encoding:
{enc}

Questions:
{questions}

Answer with JSON only: {"answers": ["...", ...]} with exactly {k} entries.

\end{memprompt}
\caption{\textbf{Blind-answerer prompt.} The verifier tests whether a reader can
recover planner facts from the encoding alone.}
\label{fig:prompt_verifier_answerer}
\end{figure*}

\begin{figure*}[!htbp]
\centering
\begin{memprompt}{ContextFolder Verifier: Fact Judge}

Reference key facts with expected answers:
{facts}

A blind reader who saw only the compressed encoding answered the verification questions as follows:
{answers}

For each item, decide whether the fact is recoverable from the encoding. Be LENIENT: accept semantically close or partial answers, accept answers that are
reasonably inferable from the encoding even if wording differs; only exact numbers/dates/names must match. Mark false only when the answer is missing, "UNKNOWN",
or contradicts the fact.
Answer with JSON only: {"ok": [true, false, ...]} with exactly {k} entries.

\end{memprompt}
\caption{\textbf{Fact-judge prompt.} The judge evaluates recoverability from the
blind reader's answers and uses lenient semantic matching.}
\label{fig:prompt_verifier_judge}
\end{figure*}

\begin{figure*}[!htbp]
\centering
\begin{memprompt}{ContextFolder Revision Feedback Prompt}
VERIFIER FEEDBACK on your previous attempt ({prev_tokens} tokens, budget {target}):
---
{prev}
---
{issues}
Re-encode by EDITING the previous encoding. Restore ONLY the facts listed as MUST-FIX; do not spend tokens on optional facts unless you are well under budget;
if over budget, tighten wording or drop optional facts. Output ONLY the new encoding.

The rule-generated issue block has one or more of the following forms:

MUST-FIX (required facts the blind reader could not recover):
- {fact}
...
Express THESE explicitly; clarity first for them.

OVER BUDGET: {ntok} vs {target} tokens. Tighten wording, merge lines, or drop the {n} optional facts entirely.

Note: {n} optional (low/inferable) facts were dropped; that is acceptable.

\end{memprompt}
\caption{\textbf{Revision feedback prompt.} The feedback is appended to later
compression rounds and contains only rule-generated revision signals.}
\label{fig:prompt_feedback}
\end{figure*}

\begin{figure*}[!htbp]
\centering
\begin{memprompt}{ContextFolder Degenerate Fallback Prompt}

You are an intelligent memory summarization assistant. Please extract the core information from the following {n} memory entries and their timestamps, and
generate a concise and coherent summary paragraph. If the memories contain relative time expressions, convert them into specific dates based on the corresponding
timestamps. The summary length should be kept within the combined length of the original memories as much as possible. Output only the final summarized result,
without any explanations.
  {lines}

\end{memprompt}
\caption{\textbf{Degenerate fallback prompt.} This one-shot summarizer handles
groups for which the planner does not return usable facts.}
\label{fig:prompt_oneshot}
\end{figure*}

\begin{figure*}[!htbp]
\centering
\begin{memprompt}{Memory Question-Answering Prompt}

You are an intelligent memory assistant tasked with retrieving accurate information from conversation memories.

# CONTEXT:
You have access to memories from two speakers in a conversation. These memories contain
timestamped information that may be relevant to answering the question.

# INSTRUCTIONS:
1. Carefully analyze all provided memories from both speakers
2. Pay special attention to the timestamps to determine the answer
3. If the question asks about a specific event or fact, look for direct evidence in the memories
4. If the memories contain contradictory information, prioritize the most recent memory
5. If there is a question about time references (like "last year", "two months ago", etc.),
   calculate the actual date based on the memory timestamp. For example, if a memory from
   4 May 2022 mentions "went to India last year," then the trip occurred in 2021.
6. Always convert relative time references to specific dates, months, or years. For example,
   convert "last year" to "2022" or "two months ago" to "March 2023" based on the memory
   timestamp. Ignore the reference while answering the question.
7. Focus only on the content of the memories from both speakers. Do not confuse character
   names mentioned in memories with the actual users who created those memories.
8. The answer should be less than 5-6 words.

# APPROACH (Think step by step):
1. First, examine all memories that contain information related to the question
2. Examine the timestamps and content of these memories carefully
3. Look for explicit mentions of dates, times, locations, or events that answer the question
4. If the answer requires calculation (e.g., converting relative time references), show your work
5. Formulate a precise, concise answer based solely on the evidence in the memories
6. Double-check that your answer directly addresses the question asked
7. Ensure your final answer is specific and avoids vague time references

Memories for user {{speaker_1_user_id}}:

{{speaker_1_memories}}

Memories for user {{speaker_2_user_id}}:

{{speaker_2_memories}}

Question: {{question}}

Answer:

\end{memprompt}
\caption{\textbf{Memory QA prompt.} The prompt answers benchmark questions from
the retrieved memories and their timestamps.}
\label{fig:prompt_qa_answer}
\end{figure*}

\begin{figure*}[!htbp]
\centering
\begin{memprompt}{Compact-Encoding Decode Legend}

# IMPORTANT - COMPACT ENCODED MEMORIES:
Some memories are stored in a compact machine-oriented encoding rather than plain
English. BEFORE answering, silently decode each memory into the facts it encodes.
Decoding legend: `@YYYY-MM-DD` = the fact happened on that date; `A -> B` = A
changed into / transferred to B; `E: x` = entity E has attribute or state x;
`~` = negation; `?` = uncertain or rumored; `;` and newlines separate facts;
compact wording drops filler words but always preserves entity names, exact
numbers, dates and relation directions. First expand each encoded line into its
plain-English facts, then answer strictly from the decoded facts.

\end{memprompt}
\caption{\textbf{Decode legend for compact memories.} This legend is injected
before the QA instructions when the retrieved records use ContextFolder encodings.}
\label{fig:prompt_qa_decode}
\end{figure*}

\begin{figure*}[!htbp]
\centering
\begin{memprompt}{Answer Accuracy Judge Prompt}

Your task is to label an answer to a question as ’CORRECT’ or ’WRONG’. You will be given the following data:
    (1) a question (posed by one user to another user),
    (2) a ’gold’ (ground truth) answer,
    (3) a generated answer
which you will score as CORRECT/WRONG.

The point of the question is to ask about something one user should know about the other user based on their prior conversations.
The gold answer will usually be a concise and short answer that includes the referenced topic, for example:
Question: Do you remember what I got the last time I went to Hawaii?
Gold answer: A shell necklace
The generated answer might be much longer, but you should be generous with your grading - as long as it touches on the same topic as the gold answer, it should
be counted as CORRECT.

For time related questions, the gold answer will be a specific date, month, year, etc. The generated answer might be much longer or use relative time
references (like "last Tuesday" or "next month"), but you should be generous with your grading - as long as it refers to the same date or time period as the gold-answer, it should
be counted as CORRECT. Even if the format differs (e.g., "May 7th" vs "7 May"), consider it CORRECT if it's the same date.

Now it's time for the real question:
Question: {question}
Gold answer: {gold_answer}
Generated answer: {generated_answer}

First, provide a short (one sentence) explanation of your reasoning, then finish with CORRECT or WRONG.
Do NOT include both CORRECT and WRONG in your response, or it will break the evaluation script.

Just return the label CORRECT or WRONG in a json format with the key as "label".

\end{memprompt}
\caption{\textbf{Answer-accuracy judge prompt.} The judge labels a generated
answer against the benchmark question and gold answer.}
\label{fig:prompt_qa_judge}
\end{figure*}

\endgroup

\end{document}